\documentclass[letterpaper]{article} 
\usepackage[preprint]{aaai2027}  
\usepackage[hyphens]{url}  
\usepackage{graphicx} 
\usepackage{natbib}  
\usepackage{caption} 
\usepackage{amsmath}
\usepackage{amssymb}
\usepackage{booktabs}
\usepackage{multirow}
\usepackage{array}

\title{TangentPDE: Learning Local Operator Responses with Sparse Finite-Difference Supervision}
\author{
Xinjian Xie\textsuperscript{\rm 1}\equalcontrib,
Ruxuan Chen\textsuperscript{\rm 2,\rm 3}\equalcontrib,
Jiashu Li\textsuperscript{\rm 4},\\
Tianyu Gao\textsuperscript{\rm 5}\corresponding,
Ye Wei\textsuperscript{\rm 5}\corresponding
}
\affiliations{
\textsuperscript{\rm 1}Sichuan Agricultural University, China\\
\textsuperscript{\rm 2}Harbin Engineering University, China\\
\textsuperscript{\rm 3}University of Southampton, United Kingdom\\
\textsuperscript{\rm 4}Wenzhou-Kean University, China\\
\textsuperscript{\rm 5}City University of Hong Kong, Hong Kong SAR, China
}

\begin{document}
\maketitle

\begin{abstract}
Neural PDE operators are often optimized through state reconstruction, which largely constrains operator values at observed inputs but does not fully characterize their sensitivity to local perturbations. A more faithful approximation of operator behavior requires not only accurate state predictions, but also correct variations along local perturbation directions. We present TangentPDE, which uses finite differences to approximate directional derivatives and constructs task-relevant finite-step responses from the state outputs of a reference solver at sparse paired inputs. Its objective jointly constrains state fidelity, response-direction alignment, and response-magnitude calibration, thereby learning both operator values and their local variations. Because the response labels are constructed solely from state outputs at paired inputs, the method requires neither explicit derivative labels, linearized or adjoint PDE solves, nor full Jacobians, and remains applicable to nonlinear solvers without derivative interfaces. Building on this, we additionally propose a selectively executed adaptation mechanism to strengthen local operator representations under finite-difference supervision. Budget-matched and correspondence-perturbed controls isolate the effect of relational supervision from improvements attributable to additional state observations or increased data volume. Designed as a plug-and-play training framework, TangentPDE integrates with diverse neural-operator backbones. Across four PDE systems and 20 task--backbone settings, TangentPDE lowers the three-seed mean state error in 18 settings and the corresponding Rollout@20 error in 18, while intrinsic finite-step response gains are strongest on DiffReact and remain task- and backbone-dependent elsewhere.
\end{abstract}

\section{Introduction}

Neural operators learn mappings between function spaces and provide fast data-driven surrogates for partial differential equations (PDEs) \citep{li2020mgno,lu2021deeponet,li2021fno,kovachki2021universal,lu2022comparison,kovachki2023neuraloperator}. Standard supervision fits a predictor $G_\theta(z)$ to a reference state $\mathcal{S}(z)$ over sampled input--output function pairs. State reconstruction remains indispensable because it controls operator values at the observed inputs and, for any fixed nonzero perturbation step, vanishing errors at both endpoints imply vanishing finite-step response error. Our focus is the finite-accuracy regime, where independent endpoint reconstruction leaves the cross-endpoint error increment implicit.

The central observation is simple but consequential. Let $e_\theta(z)=G_\theta(z)-\mathcal{S}(z)$ and let $z^+$ be a valid perturbation of $z$. Then
\[
r_{\mathrm{op}}-r^*
=
\frac{e_\theta(z^+)-e_\theta(z)}{\epsilon},
\]
where $r^*$ and $r_{\mathrm{op}}$ are the reference and learned finite-step responses. Endpoint state errors therefore control a coarse upper bound on response error, but their aggregate magnitudes do not determine the realized response mismatch. Models with comparable state error can retain different cross-endpoint error increments and hence different local operator behavior. This distinction matters whenever a learned operator is used to compare nearby inputs, estimate sensitivity, support local design or control, or advance a system repeatedly.

Derivative- and Sobolev-informed studies show that local variation supervision can improve not only derivative fidelity but also function or state approximation \citep{czarnecki2017sobolev,olearyroseberry2024dino,cao2025dino,qiu2024derivative,peyvan2025fusion}. This motivates an intermediate supervision regime: nearby admissible states can be queried from the reference system, while Jacobians, tangent solves, adjoints, or a differentiable solver interface may be unavailable or unnecessary at the task-defined perturbation scale. TangentPDE addresses this regime by converting correctly paired endpoint states into explicit finite-step relational supervision. The required reference information remains state-only, while ordinary backpropagation is used to train the neural operator.

We introduce TangentPDE-Core, which combines matched endpoint reconstruction with primary response-direction alignment and weaker response-magnitude calibration under sparse paired supervision. The novelty is not finite differencing itself, but the identification and direct optimization of the cross-endpoint error increment that remains implicit under independent state fitting. Core operates at the supervision level and preserves the backbone's state-output semantics, making it architecture-orthogonal to heterogeneous neural-operator designs. As an optional extension, the compact Layer-Selective Tangent Head (LSTH) maps finite differences of selected hidden representations to an auxiliary response residual. Its correction never enters the state outputs, so the auxiliary estimate remains distinct from the intrinsic state-output response.

Our contributions are:
\begin{itemize}
    \item an exact error-increment characterization identifying cross-endpoint state-error variation as the quantity governing finite-step response mismatch;
    \item TangentPDE-Core, a state-query-only objective that directly supervises intrinsic finite-step responses through direction alignment and weak magnitude calibration;
    \item an architecture-orthogonal evaluation framework with matched controls for endpoint observations, query budget, correspondence, and sparse pair coverage; and
    \item LSTH, a selectively executed auxiliary response learner that reuses selected hidden finite differences without changing standard single-input state inference.
\end{itemize}

\section{Related Work}

\subsection{Neural PDE Operator Learning}

DeepONet, Fourier, graph, message-passing, latent, recurrent, state-space, and attention-based neural operators mainly differ in representation, geometry handling, and information exchange \citep{li2020mgno,lu2021deeponet,li2021fno,brandstetter2022mpnpde,hao2023gnot,li2023geofno,wang2024lno,rahman2024codano,wu2024transolver,wang2025cvit,zheng2024mambano,tiwari2025lamo,ye2025rno}. Recent AAAI work further studies efficient kernels, locality, linearized physics-attention, and learning from partial observations \citep{koren2026svdno,hu2026linearno,zhou2026saot,hou2026lano}. TangentPDE is architecture-orthogonal: rather than redesigning the state predictor, it changes the supervision relation imposed on correctly paired operator evaluations while preserving the backbone's output semantics.

\subsection{Derivative-, Sensitivity-, and Sobolev-Informed Learning}

Sobolev and derivative-informed learning supervise local variation beyond function values, including target gradients, full or reduced Jacobians, Fr\'echet derivatives, Jacobian--vector products, and sensitivities obtained from tangent or adjoint systems \citep{czarnecki2017sobolev,olearyroseberry2024dino,cao2025dino,qiu2024derivative,peyvan2025fusion}. These methods directly constrain infinitesimal behavior when reliable derivative sources are available. TangentPDE occupies a distinct access regime: it learns task-scale finite-step responses from paired state queries rather than approximating a complete infinitesimal derivative operator. This replaces specialized derivative-interface requirements with ordinary state evaluations at admissible nearby inputs.

\subsection{Finite-Difference, Augmentation, and Consistency Supervision}

Finite differences also appear in data augmentation, numerical derivative approximation, and consistency objectives, but their supervision semantics differ from TangentPDE. Ordinary two-point augmentation treats the anchor and perturbed input as independent state examples and primarily enlarges input coverage. Consistency regularization commonly encourages invariance, equivariance, or agreement between related predictions. TangentPDE instead treats the anchor--perturbation correspondence itself as a supervised object and uses the reference endpoint difference as a generally nonzero target with both direction and scale. Although the target contains no reference information beyond the endpoint states, its coupled objective directly constrains the cross-endpoint error increment and is not reducible to independent endpoint reconstruction.

\subsection{Physics-Informed Correction and Auxiliary Adaptation}

Physics-informed training and rollout-stabilization methods impose governing-equation residuals, multi-step objectives, generative refinement, or inference-time numerical corrections \citep{li2024pino,lippe2023pderefiner,feng2025rplpo,huang2026physicscorrect}. TangentPDE instead supervises the finite-step behavior exhibited by the reference operator and evaluates response fidelity separately from rollout stability. LSTH is related to auxiliary multi-task learning and compact adaptation modules \citep{houlsby2019adapter}, but its adapters consume pairwise hidden finite differences rather than a single hidden state and produce a dedicated auxiliary response estimate. Standard state inference remains head-free, while intrinsic and auxiliary responses are evaluated as separate objects.

\section{Method}

\begin{figure*}[t]
\centering
\includegraphics[width=0.97\textwidth]{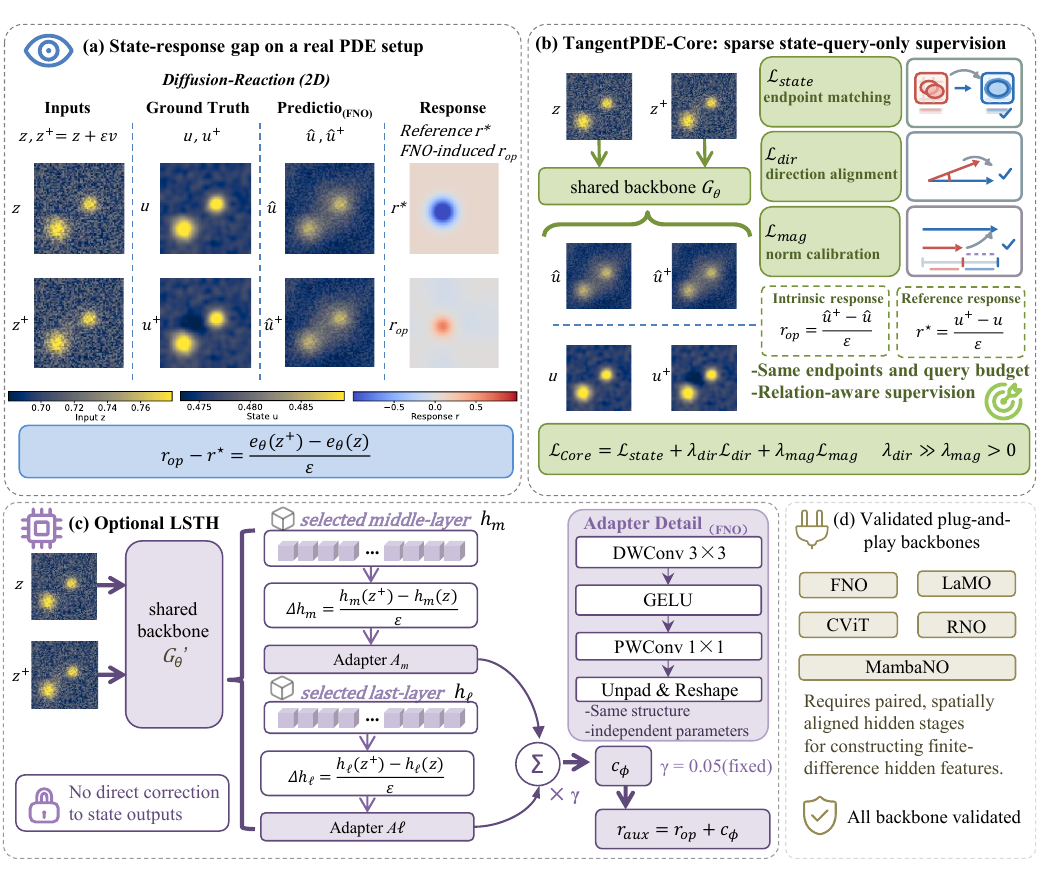}
\caption{Overview of TangentPDE. Core supervises the intrinsic state-output response $r_{\mathrm{op}}$; optional LSTH produces a separate auxiliary estimate $r_{\mathrm{aux}}$ from selected hidden finite differences.}
\label{fig:overview}
\end{figure*}

\subsection{Problem Setup and Finite-Step Responses}

Let $\mathcal{S},G_\theta:\mathcal{Z}_{\mathrm{adm}}\rightarrow\mathcal{U}$ denote the reference and learned operators. Each benchmark samples an admissible input $z$ and a perturbation protocol $(v,\epsilon)$, producing
\begin{equation}
\begin{aligned}
z^+
&=
T(z,v,\epsilon)\in\mathcal{Z}_{\mathrm{adm}},\\
T(z,v,\epsilon)
&=
z+\epsilon v
\quad\text{when no transform is required}.
\end{aligned}
\label{eq:perturbation}
\end{equation}
Ordinary and paired observations can be written compactly as
\begin{equation}
\begin{aligned}
\mathcal{D}_s&=\{(z_i,u_i):u_i=\mathcal{S}(z_i)\},\\
\mathcal{D}_p&=\{(z_j,z_j^+,u_j,u_j^+,\epsilon_j)\}.
\end{aligned}
\label{eq:datasets}
\end{equation}
For each pair, $u=\mathcal{S}(z)$ and $u^+=\mathcal{S}(z^+)$. The reference finite-step response and the intrinsic response induced by the state predictor are
\begin{equation}
r^*=\frac{u^+-u}{\epsilon},
\qquad
r_{\mathrm{op}}=
\frac{G_\theta(z^+)-G_\theta(z)}{\epsilon}.
\label{eq:responses}
\end{equation}
These finite-step secants are the actual optimization objects. They approach directional derivatives only under the corresponding differentiability and small-step conditions. We assume neither a normalized direction nor recovery of a complete Jacobian, and all locality claims refer to the benchmark-defined perturbation neighborhood. The inner product and norm below are those used by the discrete response implementation.

\subsection{Finite-Accuracy Error Increment}

With $e_\theta(z)=G_\theta(z)-\mathcal{S}(z)$,
\begin{equation}
r_{\mathrm{op}}-r^*
=
\frac{e_\theta(z^+)-e_\theta(z)}{\epsilon}.
\label{eq:error_identity}
\end{equation}
The triangle inequality gives
\begin{equation}
\lVert r_{\mathrm{op}}-r^*\rVert_2
\leq
\frac{\lVert e_\theta(z^+)\rVert_2+\lVert e_\theta(z)\rVert_2}{\epsilon}.
\label{eq:response_bound}
\end{equation}
\paragraph{Cross-endpoint error-increment characterization.}
Equation~\eqref{eq:error_identity} exactly characterizes finite-step response error through the cross-endpoint increment of state-prediction errors. Equation~\eqref{eq:response_bound} shows that endpoint accuracy limits the possible response error but does not determine the realized mismatch. In particular, endpoint-error pairs $(a,a)$ and $(a,-a)$ have identical marginal norms but induce response errors $0$ and $-2a/\epsilon$, respectively. TangentPDE-Core directly optimizes this relational quantity, which remains implicit under independent endpoint reconstruction.

\subsection{Sparse Paired Observations}

Pairs are constructed after data splitting. Their anchor labels are reused, perturbed endpoint labels are queried only for selected training anchors, and unpaired samples may continue to provide ordinary state supervision. Pair coverage is
\begin{equation}
\rho=
\frac{N_{\mathrm{paired\ anchors}}}
{N_{\mathrm{training\ anchors}}}.
\label{eq:coverage}
\end{equation}
Actual additional solver calls are reported from the pair-generation pipeline. MSE-2N-random tests whether additional state volume explains the gain; Two-point MSE matches the exact endpoint labels and query budget; Pair-shuffled Tangent tests whether preserving the implemented correspondence adds value beyond those matched observations.

\subsection{TangentPDE-Core}

For a paired sample, matched endpoint reconstruction is
\begin{equation}
\mathcal{L}_{\mathrm{state}}
=
\operatorname{MSE}(G_\theta(z),u)
+
\operatorname{MSE}(G_\theta(z^+),u^+).
\label{eq:state_loss}
\end{equation}
Two-point MSE uses the same paired endpoints and reference queries but optimizes only their marginal state errors. TangentPDE-Core additionally constrains the relation between the two predictions through
\begin{equation}
\begin{aligned}
\mathcal{L}_{\mathrm{dir}}(a,b)
&=
1-
\frac{\langle a,b\rangle}
{\lVert a\rVert_2\lVert b\rVert_2+\delta},\\
\mathcal{L}_{\mathrm{mag}}(a,b)
&=
\left|
\log(\lVert a\rVert_2+\delta)
-
\log(\lVert b\rVert_2+\delta)
\right|.
\end{aligned}
\label{eq:response_losses}
\end{equation}
Direction alignment is the primary structural term; the lower-weight magnitude term calibrates responses that are directionally correct but systematically too weak or too strong. The complete objective is
\begin{equation}
\begin{aligned}
\mathcal{L}_{\mathrm{Core}}
&=
\mathcal{L}_{\mathrm{state}}
+\lambda_{\mathrm{dir}}
\mathcal{L}_{\mathrm{dir}}(r_{\mathrm{op}},r^*)\\
&\quad+
\lambda_{\mathrm{mag}}
\mathcal{L}_{\mathrm{mag}}(r_{\mathrm{op}},r^*),\\
\lambda_{\mathrm{dir}}
&\gg
\lambda_{\mathrm{mag}}>0.
\end{aligned}
\label{eq:core_loss}
\end{equation}
The hierarchy specifies modeling priority rather than a guaranteed ordering of gradient norms. The state objective controls endpoint fidelity, while the response terms control the cross-endpoint variation of prediction errors. Direction provides the primary geometric signal, and magnitude acts as secondary scale calibration. Core and LSTH use a common valid-response convention for their response objectives; its frozen implementation is specified with the experimental protocol.

\subsection{Optional Layer-Selective Tangent Head}

Let $G_{\theta'}$ denote the state predictor after LSTH training. Its intrinsic response remains
$r_{\mathrm{op}}=[G_{\theta'}(z^+)-G_{\theta'}(z)]/\epsilon$.
For selected stages with pairwise aligned hidden representations $k\in\mathcal{K}$,
\begin{equation}
\begin{aligned}
\Delta h_k&=\frac{h_k(z^+)-h_k(z)}{\epsilon},\\
c_\phi&=\gamma\sum_{k\in\mathcal{K}}A_k(\Delta h_k),\\
r_{\mathrm{aux}}&=r_{\mathrm{op}}+c_\phi.
\end{aligned}
\label{eq:lsth_response}
\end{equation}
Each representation-specific adapter $A_k$ maps a hidden finite-step response to the output-response space. The final projections are zero-initialized and $\gamma=0.05$. The correction is not added to either state output, so $r_{\mathrm{aux}}$ is an auxiliary paired-response estimate rather than an intrinsic state-output secant.

LSTH confines response-specific parameters and computation to representation-specific depthwise/pointwise adapters and small output projections at selected stages. The adapters are bypassed during standard state and intrinsic-response inference and are activated only for auxiliary-response estimation.

In the four-block FNO instantiation, $\mathcal{K}=\{2,4\}$ under one-based indexing, with
\begin{equation}
A_k=
P_k\circ\operatorname{PWConv}_{1\times1}
\circ\operatorname{GELU}
\circ\operatorname{DWConv}_{3\times3}.
\label{eq:fno_adapter}
\end{equation}
The selected features share a spatial grid, and $P_k$ maps hidden channels to response channels. The same interface extends to token-, sequence-, graph-, and multi-resolution backbones through correspondence-compatible adapters.

LSTH reuses the Core response losses and regularizes the correction through
\begin{equation}
\mathcal{L}_{\mathrm{corr}}
=
\frac{1}{B}\sum_{i=1}^{B}
\left(
\frac{\lVert c_{\phi,i}\rVert_2}
{\lVert r_{\mathrm{op},i}\rVert_2+\delta}
\right)^2,
\label{eq:corr_loss}
\end{equation}
\begin{equation}
\begin{aligned}
\mathcal{L}_{\mathrm{adapter}}
&=
0.1\,\mathcal{L}_{\mathrm{dir}}(r_{\mathrm{aux}},r^*)\\
&\quad+
0.01\,\mathcal{L}_{\mathrm{mag}}(r_{\mathrm{aux}},r^*)\\
&\quad+
10^{-4}\mathcal{L}_{\mathrm{corr}}.
\end{aligned}
\label{eq:adapter_objective}
\end{equation}
The regularizer limits correction magnitude relative to the intrinsic response. The auxiliary estimate remains a dedicated paired-response output rather than a correction to either state prediction.

\subsection{State-Dominant Optimization and Deployment}

Stage I uses the first 20\% of LSTH continuation updates to train only the adapters. Let $\mathcal{L}_{\mathrm{aux}}$ denote $\mathcal{L}_{\mathrm{aux\text{-}response}}$. During the remaining 80\%, early stages stay frozen while selected intermediate/late stages and the state-output head receive
\begin{equation}
\begin{aligned}
g_{\mathrm{mid/late}}
&=
\nabla_{\theta_{\mathrm{mid/late}}}
\mathcal{L}_{\mathrm{state}}
+
\mu_{\mathrm{block}}
\nabla_{\theta_{\mathrm{mid/late}}}
\mathcal{L}_{\mathrm{aux}},\\
g_{\mathrm{out}}
&=
\nabla_{\theta_{\mathrm{out}}}
\mathcal{L}_{\mathrm{state}}
+
\mu_{\mathrm{head}}
\nabla_{\theta_{\mathrm{out}}}
\mathcal{L}_{\mathrm{aux}},\\
g_\phi
&=
\nabla_\phi\mathcal{L}_{\mathrm{adapter}}.
\end{aligned}
\label{eq:gradient_routing}
\end{equation}
with $\mu_{\mathrm{block}}=0.05$ and $\mu_{\mathrm{head}}=0.02$. The two coefficients ramp linearly from zero during the first 20\% of Stage-II updates. Adapter, selected shared stages, and the output head use learning-rate multipliers $1.0$, $0.02$, and $0.05$, respectively. This state-dominant parameter routing limits interference between auxiliary response learning and the primary state objective.

Standard single-input state inference requires one backbone forward, while intrinsic paired-response estimation uses two shared forwards and output differencing. Forward-path verification records zero adapter calls in both modes; auxiliary estimation alone evaluates the selected adapters. Offline pairing reuses anchor labels and queries perturbed endpoint states without differentiating through the reference solver.

\section{Experiments}

\providecommand{\best}[1]{\textbf{#1}}
\providecommand{\second}[1]{\underline{#1}}
\providecommand{\na}{\textnormal{--}}

\begin{table*}[t]
\centering
\small
\renewcommand{\arraystretch}{0.94}
\setlength{\tabcolsep}{2.8pt}
\begin{tabular}{@{}llcccccc@{}}
\toprule
\multirow{2}{*}{Task} & \multirow{2}{*}{Backbone}
& \multicolumn{3}{c}{Aug-only}
& \multicolumn{3}{c}{TangentPDE-Core} \\
\cmidrule(lr){3-5}\cmidrule(lr){6-8}
& & State $\downarrow$ & Resp. cos. $\uparrow$ & Roll.@20 $\downarrow$
  & State $\downarrow$ & Resp. cos. $\uparrow$ & Roll.@20 $\downarrow$ \\
\midrule
\multirow{5}{*}{DiffReact}
& FNO     & 0.007529 & 0.4367 & 0.00413 & \best{0.005939} & \best{0.9358} & \best{0.00336} \\
& CViT    & 0.011589 & 0.1411 & 0.00690 & \best{0.009398} & \best{0.4721} & \best{0.00590} \\
& LaMO    & 0.007906 & 0.2455 & \best{0.00344} & \best{0.007188} & \best{0.4855} & 0.00348 \\
& MambaNO & 0.012105 & 0.0722 & 0.00830 & \best{0.010433} & \best{0.9345} & \best{0.00740} \\
& RNO     & 0.008452 & 0.5431 & 0.00580 & \best{0.006316} & \best{0.9630} & \best{0.00440} \\
\midrule
\multirow{5}{*}{Kolmogorov}
& FNO     & 0.237184 & 0.7392 & 0.22747 & \best{0.237143} & \best{0.7798} & \best{0.22030} \\
& CViT    & 0.414042 & 0.3075 & 0.43090 & \best{0.353682} & \best{0.3192} & \best{0.26900} \\
& LaMO    & 0.213051 & \best{0.4187} & 0.38450 & \best{0.200136} & 0.4000 & \best{0.36410} \\
& MambaNO & 0.488124 & 0.6189 & 0.80600 & \best{0.483679} & \best{0.6493} & \best{0.79490} \\
& RNO     & 0.223250 & \best{0.6769} & 0.39700 & \best{0.179350} & 0.5074 & \best{0.33470} \\
\midrule
\multirow{5}{*}{Burgers}
& FNO     & \best{0.156602} & 0.3901 & 0.15502 & 0.157755 & \best{0.4533} & \best{0.15108} \\
& CViT    & 0.293270 & 0.2084 & 0.25000 & \best{0.265638} & \best{0.2191} & \best{0.24610} \\
& LaMO    & 0.287690 & \best{0.2064} & 0.25420 & \best{0.286906} & 0.2056 & \best{0.25270} \\
& MambaNO & 0.683363 & \best{0.1263} & 0.72760 & \best{0.635693} & 0.1077 & \best{0.68070} \\
& RNO     & 0.321615 & 0.1989 & 0.28290 & \best{0.289794} & \best{0.2288} & \best{0.25390} \\
\midrule
\multirow{5}{*}{Advection}
& FNO     & \best{0.129336} & \best{0.0899} & \best{0.03090} & 0.135071 & 0.0881 & 0.06828 \\
& CViT    & 0.546122 & \best{0.0471} & 0.06160 & \best{0.483750} & 0.0454 & \best{0.05960} \\
& LaMO    & 0.557364 & \best{0.0492} & 0.09400 & \best{0.555104} & 0.0475 & \best{0.07730} \\
& MambaNO & 0.787117 & \best{0.0455} & 0.12610 & \best{0.775974} & 0.0403 & \best{0.09050} \\
& RNO     & 0.670729 & 0.0135 & 0.65600 & \best{0.636664} & \best{0.0184} & \best{0.60530} \\
\bottomrule
\end{tabular}
\caption{Cross-system and cross-architecture results under the matched evaluation protocol. State is relative $L_2$ error, response is intrinsic finite-step response cosine, and Roll.@20 is the 20-step rollout error. Bold marks the better value within each matched Aug-only/Core pair.}
\label{tab:main}
\end{table*}

\subsection{Experimental Setup}

We evaluate four representative PDE regimes. DiffReact captures nonlinear reaction--diffusion dynamics with spatial smoothing and coupled local reactions; Kolmogorov flow represents forced nonlinear fluid dynamics with multi-scale interactions and long-horizon instability; Burgers isolates nonlinear conservation-law behavior with steep gradients and shock-like transport; and Advection isolates phase propagation and transport alignment. Together, these tasks span smoothing, multi-scale, sharp-gradient, and phase-sensitive regimes emphasized by modern scientific-ML benchmarks and operator-learning analyses \citep{takamoto2022pdebench,koehler2024apebench,wang2025hyperbolic}.

The breadth study evaluates FNO, RNO \citep{ye2025rno}, CViT \citep{wang2025cvit}, MambaNO \citep{zheng2024mambano}, and LaMO \citep{tiwari2025lamo} under matched task-specific protocols; Table~\ref{tab:budget} provides additional budget-matched controls for FNO. Core and Two-point MSE share the same $(z,z^+,u,u^+)$ pairs, splits, seeds, and endpoint states. MSE-2N matches the state count, while Pair-shuffled retains endpoint supervision but breaks correspondence. We report state relative $L_2$, response cosine, response relative $L_2$, magnitude error, and Rollout@20. Unless noted, entries are three-seed means. Per-seed results, sample variances, standard deviations, 95\% confidence intervals, effect sizes, paired tests, bootstrap analyses, and multiple-comparison-adjusted $p$-values are reported in the appendix.

\paragraph{Perturbation protocols.}
Each task uses a benchmark-specific admissible perturbation protocol. Table~\ref{tab:protocol} reports the frozen perturbation step, valid train/validation/test pair counts, and the fraction of all training anchors receiving paired supervision. Training and evaluation use disjoint sampled pairs; unseen-direction and altered-step evaluations are reported separately from the in-distribution results.

\begin{table}[tb]
\centering
\small
\renewcommand{\arraystretch}{1.04}
\setlength{\tabcolsep}{2.8pt}
\begin{tabular}{@{}llll@{}}
\toprule
Task & Step & Pairs & $\rho_{\mathrm{tr}}$ \\
\midrule
DiffReact
& $\epsilon=2.391{\times}10^{-3}$
& 200/50/100
& 100\% \\
Kolmogorov
& $\epsilon=1.0{\times}10^{-2}$
& 200/50/100
& 100\% \\
Burgers
& \shortstack[l]{$\epsilon_{\mathrm{rel}}=0.05$\\
$\bar{\epsilon}_{\mathrm{eff}}=2.387{\times}10^{-2}$}
& 1638/204/206
& 16.38\% \\
Advection
& \shortstack[l]{$\epsilon_{\mathrm{rel}}=0.05$\\
$\bar{\epsilon}_{\mathrm{eff}}=2.387{\times}10^{-2}$}
& 1638/204/206
& 16.38\% \\
\bottomrule
\end{tabular}
\caption{Benchmark-specific perturbation protocols. Pairs reports valid train/validation/test pair counts, and $\rho_{\mathrm{tr}}$ is the fraction of all training anchors receiving paired supervision. Constructing one perturbed endpoint per pair requires 350 additional reference-state queries for DiffReact and Kolmogorov and 2,048 for Burgers and Advection.}
\label{tab:protocol}
\end{table}

We organize the evaluation from breadth to mechanism. Table~\ref{tab:main} establishes transfer across systems and backbones. Table~\ref{tab:budget} controls observation count and independent two-endpoint fitting, while Table~\ref{tab:mechanism} separates direct response regression, direction--magnitude decomposition, and pair correspondence. Finally, Table~\ref{tab:lsth} evaluates LSTH across shared-state, intrinsic-response, and auxiliary-response metrics. The control tables use their respective matched protocols, and their absolute values are interpreted within each table rather than directly across tables.

\subsection{Cross-System and Cross-Architecture Results}

Table~\ref{tab:main} compares Aug-only with Core across 20 task--backbone settings. Core lowers state error and Rollout@20 in 18 settings, showing broad but non-uniform transfer across architectures and PDE regimes. Intrinsic-response gains are strongest on DiffReact and remain heterogeneous across the other systems and backbones. The aggregate pattern is therefore not a claim that response supervision uniformly improves every measured quantity; rather, it shows that paired finite-step supervision is broadly compatible with different operator parameterizations while preserving the same state-query interface. The mixed response behavior on transport-dominated systems motivates the matched controls below, which separate endpoint coverage, pair correspondence, and explicit response geometry.

\paragraph{Same-data and budget controls.}

Table~\ref{tab:budget} separates gains from additional state observations, matched endpoint fitting, and explicit finite-step response supervision. On DiffReact, MSE-2N and Two-point provide only limited improvements over the state baseline, whereas Core substantially improves all five reported metrics. This establishes a clear contribution from the relational objective beyond state count and paired endpoint reconstruction. On Kolmogorov, the three state-only or endpoint-only controls attain nearly identical one-step state errors. Core slightly increases state error but improves response cosine, response relative $L_2$, magnitude calibration, and Rollout@20, revealing a task-dependent state--response trade-off. The comparison also distinguishes the role of correspondence from the role of raw endpoint coverage: using the same two endpoint states without supervising their finite-step relation is not sufficient to reproduce the Core response gains. We therefore interpret the budget controls as a conservative check that the observed improvements are tied to relational supervision rather than to a larger state-fitting dataset.

\begin{table}[t]
\centering
\small
\renewcommand{\arraystretch}{1.00}
\setlength{\tabcolsep}{2.0pt}
\begin{tabular}{@{}lcccc@{}}
\toprule
Metric & MSE-1$\times$ & MSE-2N & Two-point & Core \\
\midrule
\multicolumn{5}{@{}l}{\textit{DiffReact}} \\
State $\downarrow$     & 0.009452 & 0.009445 & \second{0.009223} & \best{0.005908} \\
Resp. cos. $\uparrow$  & 0.2935 & 0.2944 & \second{0.2960} & \best{0.9308} \\
Resp. rel. $L_2$ $\downarrow$ & 0.9773 & 0.9768 & \second{0.9733} & \best{0.3758} \\
Mag. err. $\downarrow$ & 1.104 & 1.102 & \second{1.050} & \best{0.0997} \\
Roll.@20 $\downarrow$  & 0.004679 & 0.004687 & \second{0.004440} & \best{0.003573} \\
\midrule
\multicolumn{5}{@{}l}{\textit{Kolmogorov}} \\
State $\downarrow$     & 0.2810 & 0.2810 & 0.2810 & 0.2846 \\
Resp. cos. $\uparrow$  & 0.7204 & 0.7204 & \second{0.7205} & \best{0.7528} \\
Resp. rel. $L_2$ $\downarrow$ & 4.053 & \second{4.052} & \second{4.052} & \best{3.918} \\
Mag. err. $\downarrow$ & 1.108 & 1.108 & 1.108 & \best{1.095} \\
Roll.@20 $\downarrow$  & 0.2960 & 0.2960 & \second{0.2959} & \best{0.2945} \\
\bottomrule
\end{tabular}
\caption{Budget-matched FNO controls. MSE-2N matches the state count; Two-point uses the same paired endpoints and reference-query budget as Core but omits response supervision.}
\label{tab:budget}
\end{table}

\paragraph{Objective and correspondence analysis.}

Table~\ref{tab:mechanism} separates direct response regression, the direction--magnitude decomposition, and pair correspondence. State + Direct Resp. MSE is the undecomposed response-learning baseline and fits the full response vector with a single Euclidean objective. On DiffReact, direct response regression improves over state-only continuation but remains far below direction supervision in response cosine; Full Core gives the lowest state error while matching the best rollout. On Kolmogorov, Direct Resp. MSE gives the lowest state and rollout errors, whereas the response variants remain below the state-only response cosine. The pair controls further show that exact correspondence has clear marginal value on DiffReact but only a small incremental effect over the paired state-preserving protocol on Kolmogorov.

\begin{table}[t]
\centering
\small
\renewcommand{\arraystretch}{1.00}
\setlength{\tabcolsep}{2.1pt}
\begin{tabular}{@{}lccc@{}}
\toprule
Setting & State $\downarrow$ & Resp. cos. $\uparrow$ & Roll.@20 $\downarrow$ \\
\midrule
\multicolumn{4}{@{}l}{\textit{Objective composition: DiffReact}} \\
State only   & 0.008536 & 0.5360 & 0.0059 \\
State + Direct Resp. MSE & 0.008095 & 0.5953 & \second{0.0050} \\
State + Dir. & \second{0.007107} & \best{0.9640} & \best{0.0044} \\
State + Mag. & 0.008715 & 0.4311 & 0.0065 \\
Full Core    & \best{0.006316} & \second{0.9630} & \best{0.0044} \\
\midrule
\multicolumn{4}{@{}l}{\textit{Objective composition: Kolmogorov}} \\
State only   & 0.223495 & \best{0.6775} & 0.3959 \\
State + Direct Resp. MSE & \best{0.179553} & 0.5073 & \best{0.3350} \\
State + Dir. & \second{0.179643} & 0.5089 & \second{0.3351} \\
State + Mag. & 0.179648 & 0.5077 & \second{0.3351} \\
Full Core    & 0.179683 & \second{0.5090} & \second{0.3351} \\
\midrule
\multicolumn{4}{@{}l}{\textit{Pair correspondence: DiffReact}} \\
State baseline & 0.008267 & 0.5349 & 0.0057 \\
Pair-shuffled  & \second{0.006797} & \second{0.9297} & \second{0.0050} \\
True Core      & \best{0.006316} & \best{0.9630} & \best{0.0044} \\
\midrule
\multicolumn{4}{@{}l}{\textit{Pair correspondence: Kolmogorov}} \\
State baseline & 0.222942 & \best{0.6783} & 0.3936 \\
Pair-shuffled  & \second{0.179625} & 0.5061 & \best{0.3350} \\
True Core      & \best{0.179574} & \second{0.5068} & \best{0.3350} \\
\bottomrule
\end{tabular}
\caption{RNO mechanism controls. Objective rows isolate response-regression, direction, and magnitude terms; pair rows break or retain endpoint correspondence.}
\label{tab:mechanism}
\end{table}

\paragraph{Sparse paired supervision.}

Figure~\ref{fig:sparse} jointly evaluates state and intrinsic-response pair efficiency on DiffReact. Two-point MSE changes only modestly with the paired-anchor budget. In contrast, TangentPDE-Core rapidly improves intrinsic response alignment and continues to reduce state error as more paired anchors are introduced. The distinct response and state trends show that the relational objective affects both local output variation and endpoint prediction, with different sensitivities to pair coverage. The response curve improves most sharply at small budgets, indicating that a limited number of paired secants can already provide a useful local direction signal. The state curve changes more gradually, which is consistent with state prediction depending on both endpoint coverage and the regularization induced by the response objective.

\begin{center}
\begin{minipage}{\columnwidth}
\centering
\includegraphics[width=0.82\columnwidth]{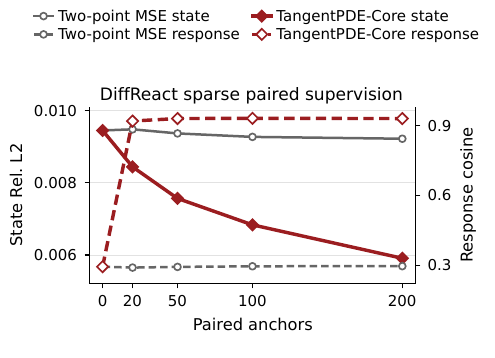}
\captionof{figure}{Pair efficiency on DiffReact. State relative $L_2$ error and intrinsic finite-step response cosine are reported as paired training anchors increase. Two-point MSE fits endpoint states independently, whereas TangentPDE-Core also supervises their finite-step relation.}
\label{fig:sparse}
\end{minipage}
\end{center}

\paragraph{Layer-selective tangent head.}

Table~\ref{tab:lsth} separates changes in the shared state predictor from gains of the dedicated auxiliary response path. LSTH markedly improves auxiliary response alignment, while its intrinsic response remains task dependent. This confirms that the head primarily provides a specialized paired-response estimate rather than uniformly modifying the state operator's intrinsic finite-step response.

\begin{table}[t]
\centering
\small
\renewcommand{\arraystretch}{1.00}
\setlength{\tabcolsep}{1.35pt}
\begin{tabular}{@{}lcccc@{}}
\toprule
Variant & State $\downarrow$ & Intr. resp. $\uparrow$ & Aux. resp. $\uparrow$ & Roll. $\downarrow$ \\
\midrule
\multicolumn{5}{@{}l}{\textit{DiffReact}} \\
Static   & 0.0059 & \second{0.9433} & \na & 0.0035 \\
State FT & \best{0.0032} & 0.9410 & \na & \best{0.0018} \\
Weak FT  & 0.0037 & \best{0.9440} & \na & 0.0021 \\
LSTH     & \second{0.0036} & 0.9431 & 0.9779 & \second{0.0019} \\
\midrule
\multicolumn{5}{@{}l}{\textit{Kolmogorov}} \\
Static   & 0.2809 & \best{0.7572} & \na & 0.2941 \\
State FT & \best{0.2749} & 0.7281 & \na & 0.2887 \\
Weak FT  & \second{0.2795} & \second{0.7525} & \na & \second{0.2886} \\
LSTH     & \best{0.2749} & 0.7281 & 0.9071 & \best{0.2885} \\
\bottomrule
\end{tabular}
\caption{Layer-selective response-aware fine-tuning. Entries are three-seed means; standard deviations and confidence intervals are reported in the appendix. Intrinsic response is obtained from the shared state predictor for every variant, while auxiliary response is defined only for LSTH.}
\label{tab:lsth}
\end{table}

Across the two FNO configurations, LSTH adds 8.77--26.52\% parameters, while auxiliary-response evaluation adds 12.44--16.21\% latency over intrinsic paired-response inference.

\begin{center}
\begin{minipage}{\columnwidth}
\centering
\includegraphics[width=0.92\columnwidth]{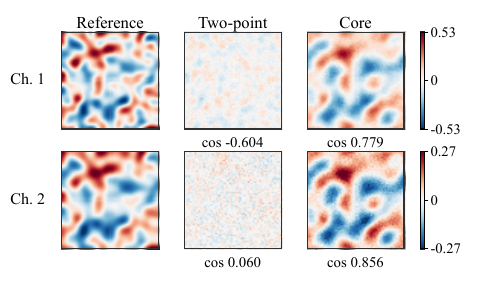}
\captionof{figure}{Response-field comparison for a representative paired query. Reference finite-step responses are shown against Two-point MSE and TangentPDE-Core predictions for the two output channels; cosine values are computed against the corresponding reference response field. This qualitative example complements the aggregate mechanism metrics in Table~\ref{tab:mechanism}.}
\label{fig:response_fields}
\end{minipage}
\end{center}

\section{Discussion and Limitations}

\paragraph{What TangentPDE learns.}

Taken together, the experiments establish TangentPDE as a finite-step response-learning framework whose relational supervision also regularizes state learning. Core improves state fidelity and rollout broadly, while the matched controls isolate contributions beyond additional state observations and independent endpoint reconstruction. Explicit response geometry provides the clearest additional value on sensitivity-relevant systems, and the response-MSE comparison shows when the direction--magnitude decomposition improves over undecomposed vector regression.

\paragraph{Task-dependent response behavior.}

Response supervision exhibits distinct behavior across the evaluated PDE systems. DiffReact shows the strongest and most consistent intrinsic-response gains, whereas Burgers and Advection retain state or rollout improvements in many settings but show weaker or architecture-dependent response changes. On Kolmogorov, the balance among state, intrinsic response, and rollout also varies across backbones. These results motivate task- and architecture-specific validation of response objectives.

\paragraph{LSTH semantics and deployment.}

LSTH improves a dedicated auxiliary response estimate by reusing selected hidden finite differences, while the intrinsic response remains defined by the shared state predictor's output difference. The two quantities are therefore reported separately. The adapters are bypassed during standard state and intrinsic-response inference and execute only for the auxiliary estimate. Full correction remains task dependent, and transport settings can over-correct when the auxiliary residual dominates the intrinsic response.

\paragraph{Scope, solver interface, and reporting.}

TangentPDE learns finite-step secants along sampled perturbation families rather than a complete Jacobian. The state-query-only interface broadens applicability to solvers without derivative, tangent, or adjoint APIs, but wall-clock cost relative to JVP, tangent, or adjoint data generation remains solver dependent. When anchor labels are reused, each new pair typically requires one additional perturbed-state solve; the exact call count and measured runtime are reported from the pair-generation manifests. The appendix provides per-seed values, variances, confidence intervals, effect sizes, corrected significance tests, response masks, perturbation protocols, hyperparameters, and solver-query accounting. Claims about unseen directions and altered step sizes are restricted to the evaluated perturbation families.

\section{Conclusion}

TangentPDE identifies the cross-endpoint error increment as the missing relational quantity in finite-accuracy state fitting and turns paired state queries into direct finite-step response supervision. TangentPDE-Core preserves the state-prediction interface while improving state fidelity and rollout across diverse PDE regimes and backbones, and controlled comparisons isolate the value of finite-step geometry beyond additional endpoint observations. The direction--magnitude decomposition separates geometric alignment from scale calibration, while LSTH extends the framework with a dedicated auxiliary response learner whose output remains distinct from the intrinsic state-output response. Together, these results establish paired finite-step response learning as a practical supervision regime between state-only fitting and explicit derivative-informed training.

\bibliography{reference}

@article{kovachki2023neuraloperator,
  title={Neural Operator: Learning Maps Between Function Spaces with Applications to PDEs},
  author={Kovachki, Nikola and Li, Zongyi and Liu, Burigede and Azizzadenesheli, Kamyar and Bhattacharya, Kaushik and Stuart, Andrew and Anandkumar, Anima},
  journal={Journal of Machine Learning Research},
  volume={24},
  number={89},
  pages={1--97},
  year={2023}
}

@inproceedings{li2021fno,
  title={Fourier Neural Operator for Parametric Partial Differential Equations},
  author={Li, Zongyi and Kovachki, Nikola and Azizzadenesheli, Kamyar and Liu, Burigede and Bhattacharya, Kaushik and Stuart, Andrew and Anandkumar, Anima},
  booktitle={International Conference on Learning Representations},
  year={2021}
}

@article{lu2021deeponet,
  title={Learning Nonlinear Operators via DeepONet Based on the Universal Approximation Theorem of Operators},
  author={Lu, Lu and Jin, Pengzhan and Pang, Guofei and Zhang, Zhongqiang and Karniadakis, George Em},
  journal={Nature Machine Intelligence},
  volume={3},
  pages={218--229},
  year={2021}
}

@article{cao2025dino,
  title={Derivative-Informed Neural Operator Acceleration of Geometric MCMC for Infinite-Dimensional Bayesian Inverse Problems},
  author={Cao, Lianghao and O'Leary-Roseberry, Thomas and Ghattas, Omar},
  journal={Journal of Machine Learning Research},
  volume={26},
  number={78},
  pages={1--68},
  year={2025}
}

@inproceedings{qiu2024derivative,
  title={Derivative-Enhanced Deep Operator Network},
  author={Qiu, Yating and Bridges, Nathan and Chen, Peng},
  booktitle={Advances in Neural Information Processing Systems},
  volume={37},
  pages={20945--20981},
  year={2024}
}

@inproceedings{houlsby2019adapter,
  title={Parameter-Efficient Transfer Learning for NLP},
  author={Houlsby, Neil and Giurgiu, Andrei and Jastrzebski, Stanislaw and Morrone, Bruna and de Laroussilhe, Quentin and Gesmundo, Andrea and Attariyan, Mona and Gelly, Sylvain},
  booktitle={International Conference on Machine Learning},
  pages={2790--2799},
  year={2019}
}

@inproceedings{hou2026lano,
  title={Learning Neural Operators from Partial Observations via Latent Autoregressive Modeling},
  author={Hou, Jingren and Wang, Hong and Xu, Pengyu and Gao, Chang and Liu, Huafeng and Jing, Liping},
  booktitle={Proceedings of the AAAI Conference on Artificial Intelligence},
  volume={40},
  pages={390--398},
  year={2026},
  doi={10.1609/aaai.v40i1.37001}
}

@inproceedings{koren2026svdno,
  title={{SVD-NO}: Learning {PDE} Solution Operators with {SVD} Integral Kernels},
  author={Koren, Noam and Mackenbach, Ralf J. J. and van Sloun, Ruud J. G. and Radinsky, Kira and Freedman, Daniel},
  booktitle={Proceedings of the AAAI Conference on Artificial Intelligence},
  volume={40},
  pages={525--533},
  year={2026},
  doi={10.1609/aaai.v40i1.37016}
}

@inproceedings{hu2026linearno,
  title={Transolver Is a Linear Transformer: Revisiting Physics-Attention Through the Lens of Linear Attention},
  author={Hu, Wenjie and Liu, Sidun and Qiao, Peng and Sun, Zhenglun and Dou, Yong},
  booktitle={Proceedings of the AAAI Conference on Artificial Intelligence},
  volume={40},
  pages={408--416},
  year={2026},
  doi={10.1609/aaai.v40i1.37003}
}

@inproceedings{zhou2026saot,
  title={{SAOT}: An Enhanced Locality-Aware Spectral Transformer for Solving {PDE}s},
  author={Zhou, Chenhong and Chen, Jie and Yang, Zaifeng},
  booktitle={Proceedings of the AAAI Conference on Artificial Intelligence},
  volume={40},
  pages={28928--28936},
  year={2026},
  doi={10.1609/aaai.v40i34.40128}
}

@inproceedings{feng2025rplpo,
  title={How to Re-enable {PDE} Loss for Physical Systems Modeling Under Partial Observation},
  author={Feng, Haodong and Wang, Yue and Fan, Dixia},
  booktitle={Proceedings of the AAAI Conference on Artificial Intelligence},
  volume={39},
  pages={182--190},
  year={2025},
  doi={10.1609/aaai.v39i1.31994}
}

@inproceedings{huang2026physicscorrect,
  title={{PhysicsCorrect}: A Training-Free Approach for Stable Neural {PDE} Simulations},
  author={Huang, Xinquan and Perdikaris, Paris},
  booktitle={Proceedings of the AAAI Conference on Artificial Intelligence},
  volume={40},
  pages={22057--22065},
  year={2026},
  doi={10.1609/aaai.v40i26.39360}
}

@inproceedings{czarnecki2017sobolev,
  title={Sobolev Training for Neural Networks},
  author={Czarnecki, Wojciech M. and Osindero, Simon and Jaderberg, Max and Swirszcz, Grzegorz and Pascanu, Razvan},
  booktitle={Advances in Neural Information Processing Systems},
  volume={30},
  year={2017}
}

@article{olearyroseberry2024dino,
  title={Derivative-Informed Neural Operator: An Efficient Framework for High-Dimensional Parametric Derivative Learning},
  author={O'Leary-Roseberry, Thomas and Chen, Peng and Villa, Umberto and Ghattas, Omar},
  journal={Journal of Computational Physics},
  volume={496},
  pages={112555},
  year={2024},
  doi={10.1016/j.jcp.2023.112555}
}

@inproceedings{takamoto2022pdebench,
  title={{PDEBench}: An Extensive Benchmark for Scientific Machine Learning},
  author={Takamoto, Makoto and Praditia, Timothy and Leiteritz, Raphael and MacKinlay, Daniel and Alesiani, Francesco and Pflueger, Dirk and Niepert, Mathias},
  booktitle={Advances in Neural Information Processing Systems},
  volume={35},
  pages={1596--1611},
  year={2022},
  doi={10.52202/068431-0117}
}

@inproceedings{zheng2024mambano,
  title={Alias-Free Mamba Neural Operator},
  author={Zheng, Jianwei and Li, Wei and Xu, Ni and Zhu, Junwei and Lin, Xiaoxu and Zhang, Xiaoqin},
  booktitle={Advances in Neural Information Processing Systems},
  volume={37},
  pages={52962--52995},
  year={2024},
  doi={10.52202/079017-1678}
}

@inproceedings{wang2025cvit,
  title={{CViT}: Continuous Vision Transformer for Operator Learning},
  author={Wang, Sifan and Seidman, Jacob H. and Sankaran, Shyam and Wang, Hanwen and Pappas, George J. and Perdikaris, Paris},
  booktitle={The Thirteenth International Conference on Learning Representations},
  year={2025},
  url={https://openreview.net/forum?id=cRnCcuLvyr}
}

@inproceedings{rahman2024codano,
  title={Pretraining Codomain Attention Neural Operators for Solving Multiphysics {PDE}s},
  author={Rahman, Md Ashiqur and George, Robert Joseph and Elleithy, Mogab and Leibovici, Daniel and Li, Zongyi and Bonev, Boris and White, Colin and Berner, Julius and Yeh, Raymond A. and Kossaifi, Jean and Azizzadenesheli, Kamyar and Anandkumar, Anima},
  booktitle={Advances in Neural Information Processing Systems},
  volume={37},
  year={2024},
  doi={10.52202/079017-3306}
}

@inproceedings{koehler2024apebench,
  title={{APEBench}: A Benchmark for Autoregressive Neural Emulators of {PDE}s},
  author={Koehler, Felix and Niedermayr, Simon and Westermann, Ruediger and Thuerey, Nils},
  booktitle={Advances in Neural Information Processing Systems},
  volume={37},
  year={2024},
  doi={10.52202/079017-3822}
}

@inproceedings{li2020mgno,
  title={Multipole Graph Neural Operator for Parametric Partial Differential Equations},
  author={Li, Zongyi and Kovachki, Nikola and Azizzadenesheli, Kamyar and Liu, Burigede and Bhattacharya, Kaushik and Stuart, Andrew and Anandkumar, Anima},
  booktitle={Advances in Neural Information Processing Systems},
  volume={33},
  pages={6755--6766},
  year={2020}
}

@article{kovachki2021universal,
  title={On Universal Approximation and Error Bounds for Fourier Neural Operators},
  author={Kovachki, Nikola and Lanthaler, Samuel and Mishra, Siddhartha},
  journal={Journal of Machine Learning Research},
  volume={22},
  number={290},
  pages={1--76},
  year={2021}
}

@article{lu2022comparison,
  title={A Comprehensive and Fair Comparison of Two Neural Operators (with Practical Extensions) Based on {FAIR} Data},
  author={Lu, Lu and Meng, Xuhui and Cai, Shengze and Mao, Zhiping and Goswami, Somdatta and Zhang, Zhongqiang and Karniadakis, George Em},
  journal={Computer Methods in Applied Mechanics and Engineering},
  volume={393},
  pages={114778},
  year={2022},
  doi={10.1016/j.cma.2022.114778}
}

@inproceedings{brandstetter2022mpnpde,
  title={Message Passing Neural {PDE} Solvers},
  author={Brandstetter, Johannes and Worrall, Daniel E. and Welling, Max},
  booktitle={International Conference on Learning Representations},
  year={2022}
}

@inproceedings{hao2023gnot,
  title={{GNOT}: A General Neural Operator Transformer for Operator Learning},
  author={Hao, Zhongkai and Wang, Zhengyi and Su, Hang and Ying, Chengyang and Dong, Yinpeng and Liu, Songming and Cheng, Ze and Song, Jian and Zhu, Jun},
  booktitle={Proceedings of the 40th International Conference on Machine Learning},
  series={Proceedings of Machine Learning Research},
  volume={202},
  pages={12556--12569},
  year={2023},
  publisher={PMLR}
}

@article{li2023geofno,
  title={Fourier Neural Operator with Learned Deformations for {PDE}s on General Geometries},
  author={Li, Zongyi and Huang, Daniel Zhengyu and Liu, Burigede and Anandkumar, Anima},
  journal={Journal of Machine Learning Research},
  volume={24},
  number={388},
  pages={1--26},
  year={2023}
}

@inproceedings{wu2024transolver,
  title={Transolver: A Fast Transformer Solver for {PDE}s on General Geometries},
  author={Wu, Haixu and Luo, Huakun and Wang, Haowen and Wang, Jianmin and Long, Mingsheng},
  booktitle={Proceedings of the 41st International Conference on Machine Learning},
  series={Proceedings of Machine Learning Research},
  volume={235},
  pages={53681--53705},
  year={2024},
  publisher={PMLR}
}

@inproceedings{wang2024lno,
  title={Latent Neural Operator for Solving Forward and Inverse {PDE} Problems},
  author={Wang, Tian and Wang, Chuang},
  booktitle={Advances in Neural Information Processing Systems},
  volume={37},
  year={2024}
}

@article{li2024pino,
  title={Physics-Informed Neural Operator for Learning Partial Differential Equations},
  author={Li, Zongyi and Zheng, Hongkai and Kovachki, Nikola and Jin, David and Chen, Haoxuan and Liu, Burigede and Azizzadenesheli, Kamyar and Anandkumar, Anima},
  journal={ACM/IMS Journal of Data Science},
  volume={1},
  number={3},
  pages={1--27},
  year={2024}
}

@inproceedings{lippe2023pderefiner,
  title={{PDE}-Refiner: Achieving Accurate Long Rollouts with Neural {PDE} Solvers},
  author={Lippe, Phillip and Veeling, Bastiaan S. and Perdikaris, Paris and Turner, Richard E. and Brandstetter, Johannes},
  booktitle={Advances in Neural Information Processing Systems},
  volume={36},
  pages={67398--67433},
  year={2023}
}

@inproceedings{tiwari2025lamo,
  title={Latent Mamba Operator for Partial Differential Equations},
  author={Tiwari, Karn and Dutta, Niladri and Krishnan, N. M. Anoop and A. P., Prathosh},
  booktitle={Proceedings of the 42nd International Conference on Machine Learning},
  series={Proceedings of Machine Learning Research},
  volume={267},
  pages={59639--59667},
  year={2025},
  publisher={PMLR}
}

@article{ye2025rno,
  title={Recurrent Neural Operators: Stable Long-Term {PDE} Prediction},
  author={Ye, Zaijun and Zhang, Chen-Song and Wang, Wansheng},
  journal={arXiv preprint arXiv:2505.20721},
  year={2025}
}

@article{wang2025hyperbolic,
  title={Operator Learning for Hyperbolic {PDE}s},
  author={Wang, Christopher and Townsend, Alex},
  journal={Journal of Machine Learning Research},
  volume={26},
  number={199},
  pages={1--44},
  year={2025}
}

@article{peyvan2025fusion,
  title={Fusion-{DeepONet}: A Data-Efficient Neural Operator for Geometry-Dependent Hypersonic and Supersonic Flows},
  author={Peyvan, Ahmad and Kumar, Varun and Karniadakis, George Em},
  journal={arXiv preprint arXiv:2501.01934},
  year={2025}
}

\clearpage
\appendix
\setcounter{secnumdepth}{2}
\setcounter{table}{0}
\setcounter{figure}{0}
\renewcommand{\thetable}{A\arabic{table}}
\renewcommand{\thefigure}{A\arabic{figure}}
\setcounter{topnumber}{4}
\setcounter{bottomnumber}{2}
\setcounter{totalnumber}{6}
\setcounter{dbltopnumber}{4}
\renewcommand{\topfraction}{0.92}
\renewcommand{\bottomfraction}{0.85}
\renewcommand{\textfraction}{0.08}
\renewcommand{\floatpagefraction}{0.80}
\renewcommand{\dbltopfraction}{0.92}
\renewcommand{\dblfloatpagefraction}{0.80}
\section{Reproducibility and Evidence Scope}

\makeatletter
\setlength{\@fptop}{0pt}
\setlength{\@fpsep}{10pt plus 2pt}
\setlength{\@fpbot}{0pt plus 1fil}
\setlength{\@dblfptop}{0pt}
\setlength{\@dblfpsep}{10pt plus 2pt}
\setlength{\@dblfpbot}{0pt plus 1fil}
\makeatother

This supplement distinguishes three evidence families that serve different
purposes. First, the submitted task-specific results reproduce the point
estimates used in the main paper. Second, independent uniform-protocol reruns
test whether the qualitative behavior persists under one common training
schedule, but they are not used to reconstruct uncertainty around the submitted
point estimates. Third, matched controls and diagnostic studies isolate
specific competing explanations or probe limited robustness properties. These
evidence families are reported separately and should not be treated as
interchangeable estimates of the same quantity.

All reported results use three random seeds, 1234, 2025, and 3407, unless
stated otherwise. The controlled FNO and RNO experiments use Adam for 50
epochs with batch size 4 and an initial learning rate of $10^{-3}$. Model
selection minimizes validation state relative $L_2$ error; test metrics are
computed once from the selected checkpoint. TangentPDE-Core uses
$\lambda_{\mathrm{dir}}=0.1$, $\lambda_{\mathrm{mag}}=0.01$, and
$\delta=10^{-8}$, matching the frozen main-text objective. These controlled runs use zero
weight decay,
full-precision training, no learning-rate scheduler, and no gradient clipping.
Loss warmup is disabled for Core and its matched controls. Any independently
rerun protocol that uses a different continuation schedule is identified
explicitly in its own subsection.

The breadth study compares FNO, CViT, LaMO, MambaNO, and RNO using the same
task split and evaluation interface for Aug-only and TangentPDE-Core.
Within each task--backbone pair, the two methods use the same state-predictor
architecture and differ only in the training objective. Core and Two-point MSE
receive identical paired endpoints. MSE-2N matches the number of state
observations, while the pair-shuffled control retains the endpoint-state
multiset and optimization budget but disrupts the declared endpoint
correspondence.

Unless a table caption states otherwise, boldface identifies the better point
estimate within the declared comparison and does not by itself imply
statistical significance. A dash denotes an undefined or inapplicable
quantity rather than a numerical zero.

Tables~\ref{tab:app_backbone_architecture} and
\ref{tab:app_backbone_training} give the architecture and optimization details
for the uniform-protocol breadth study. The task-dependent output dimension is
kept explicit because each model predicts the complete trajectory rather than
a single shared-size output tensor.

\begin{table*}[t]
\centering
\small
\setlength{\tabcolsep}{3.2pt}
\caption{Backbone configurations for the uniform-protocol breadth study. All
models use one input anchor, denoted by $C_z$ channels, and predict the complete
$T C_u$-channel trajectory. For DiffReact, $(C_z,T,C_u)=(2,51,2)$; for
Kolmogorov, $(C_z,T,C_u)=(1,21,1)$. Parameter counts (in millions) are reported
separately for these two output dimensions as D/K.}
\label{tab:app_backbone_architecture}
\begin{tabular}{lccccp{1.42in}ccr}
\toprule
Backbone & Input & Output & Blocks & Width & Backbone-specific setting & History & Prediction & Params. D/K (M) \\
\midrule
FNO     & $C_z$ & $TC_u$ & 3 & 32 & Fourier modes $8\times8$ & 1 & Parallel & 0.141/0.135 \\
CViT    & $C_z$ & $TC_u$ & 3 & 32 & Patch $4\times4$; 4 heads & 1 & Parallel & 0.043/0.039 \\
LaMO    & $C_z$ & $TC_u$ & 3 & 32 & Latent width 32; stride 2 & 1 & Parallel & 0.078/0.075 \\
MambaNO & $C_z$ & $TC_u$ & 3 & 32 & State size 16; convolution width 4 & 1 & Parallel & 0.014/0.011 \\
RNO     & $C_z$ & $TC_u$ & 3 & 32 & Three ConvGRU recurrent blocks & 1 & Recurrent & 0.176/0.176 \\
\bottomrule
\end{tabular}
\end{table*}

\begin{table*}[t]
\centering
\small
\setlength{\tabcolsep}{3.5pt}
\caption{Training configuration for the uniform-protocol breadth study. The
same optimization and checkpoint-selection settings are used for Aug-only and
TangentPDE-Core. ``None'' means that the corresponding operation is disabled.}
\label{tab:app_backbone_training}
\begin{tabular}{lccccccccp{1.18in}}
\toprule
Backbone & Optimizer & LR & Batch & Epochs & Scheduler & Weight decay & Grad. clip & Precision & Checkpoint selection \\
\midrule
FNO     & Adam & $10^{-3}$ & 4 & 50 & None & 0 & None & FP32 & Lowest validation state relative $L_2$ \\
CViT    & Adam & $10^{-3}$ & 4 & 50 & None & 0 & None & FP32 & Lowest validation state relative $L_2$ \\
LaMO    & Adam & $10^{-3}$ & 4 & 50 & None & 0 & None & FP32 & Lowest validation state relative $L_2$ \\
MambaNO & Adam & $10^{-3}$ & 4 & 50 & None & 0 & None & FP32 & Lowest validation state relative $L_2$ \\
RNO     & Adam & $10^{-3}$ & 4 & 50 & None & 0 & None & FP32 & Lowest validation state relative $L_2$ \\
\bottomrule
\end{tabular}
\end{table*}

\section{Datasets and Perturbation Protocols}

Pairs are constructed only after the anchor split, so training, validation,
and test anchors remain disjoint before any perturbation is generated. This
ordering prevents a perturbed endpoint derived from one split from being used
as an anchor or label in another split. For each valid anchor $z$, a perturbation protocol $(v,\epsilon)$
defines an admissible perturbed endpoint
$z^+=T(z,v,\epsilon)\in\mathcal Z_{\mathrm{adm}}$. Following the frozen
main-text definition, $T(z,v,\epsilon)=z+\epsilon v$ when no task-specific
transform is required. The reference solver is queried at $z^+$ while the
existing label at $z$ is reused. Thus, one accepted pair adds exactly one
perturbed-endpoint query rather than two new solver calls.
Table~\ref{tab:app_protocol} separates training-pair coverage from coverage
over all available anchors.

DiffReact uses coupled reaction--diffusion perturbations at
$\epsilon=2.391\times10^{-3}$, and Kolmogorov uses admissible flow
perturbations at $\epsilon=10^{-2}$. Burgers and Advection use normalized
random Fourier perturbations with relative step $0.05$; their mean effective
step is $2.387\times10^{-2}$. These task-specific constructions define the
finite-step response families evaluated in this work; conclusions about
unseen directions or altered step sizes are restricted to the dedicated
studies reported later.

\section{Evaluation Metrics and Statistical Reporting}

Let $G$ denote the learned state operator and let $(z,z^+,u,u^+)$ be a paired
query. The intrinsic predicted and reference responses are
\begin{equation}
 r_{\mathrm{op}}=\frac{G(z^+)-G(z)}{\epsilon},\qquad
 r^\star=\frac{u^+-u}{\epsilon}.
\end{equation}
The term \emph{intrinsic} refers to the response induced by the shared state
predictor itself; it is distinct from the auxiliary response estimate
introduced later for LSTH. These quantities are finite-step secants under the
benchmark-defined perturbation protocol, not estimates of a complete Jacobian.

For clarity, the frozen TangentPDE-Core training objective is reproduced here:
\begin{align}
 \mathcal L_{\mathrm{state}}
 &= \operatorname{MSE}(G_\theta(z),u)
  + \operatorname{MSE}(G_\theta(z^+),u^+),\\
 \mathcal L_{\mathrm{dir}}(a,b)
 &= 1-\frac{\langle a,b\rangle}
 {\lVert a\rVert_2\lVert b\rVert_2+\delta},\\
 \mathcal L_{\mathrm{mag}}(a,b)
 &= \left|\log(\lVert a\rVert_2+\delta)
 -\log(\lVert b\rVert_2+\delta)\right|,\\
 \mathcal L_{\mathrm{Core}}
 &= \mathcal L_{\mathrm{state}}
 +\lambda_{\mathrm{dir}}\mathcal L_{\mathrm{dir}}(r_{\mathrm{op}},r^\star)\notag\\
 &\quad +\lambda_{\mathrm{mag}}
 \mathcal L_{\mathrm{mag}}(r_{\mathrm{op}},r^\star).
\end{align}
The quantities below are reported evaluation metrics. They should not be
conflated with the training losses even when they use the same response pair.

State error is the sample-wise relative norm
\begin{equation}
 E_{\mathrm{state}}=
 \frac{\lVert G(z)-u\rVert_2}{\lVert u\rVert_2+\delta}.
\end{equation}
Response cosine, relative response error, and magnitude error are
\begin{align}
 C_{\mathrm{resp}} & =
 \frac{\langle r_{\mathrm{op}},r^\star\rangle}
 {\lVert r_{\mathrm{op}}\rVert_2\lVert r^\star\rVert_2+\delta},\\
 E_{\mathrm{resp}} & =
 \frac{\lVert r_{\mathrm{op}}-r^\star\rVert_2}
 {\lVert r^\star\rVert_2+\delta},\\
 E_{\mathrm{mag}} & =
 \left|\log(\lVert r_{\mathrm{op}}\rVert_2+\delta)
 -\log(\lVert r^\star\rVert_2+\delta)\right|.
\end{align}
Metrics are computed per sample and then averaged. Near-zero reference
responses are excluded from directional metrics using the response-norm
threshold $\delta$. Rollout@20 is the state relative $L_2$ error after 20
autoregressive prediction steps. Where uncertainty is shown, entries are the
mean and sample standard deviation across the three seeds.

Seed-level standard deviations describe variation due to training
randomness under a fixed configuration; they are not interpreted as
independent test-sample uncertainty. The across-setting analysis reported
later instead treats a complete task--backbone setting as the paired
resampling and testing unit. These two uncertainty summaries answer different
questions and are therefore reported separately.

Table~\ref{tab:response_mask_statistics} reports the realized mask statistics
on the frozen test pairs. No test response is removed at the fixed threshold.
The mask is applied only to response cosine; response relative $L_2$ and
magnitude error remain defined through the fixed stabilizer $\delta$ and are
not removed by this directional mask. State and rollout metrics do not use this mask.

\begin{table*}[t]
\centering
\small
\begin{tabular}{lrrrrrrr}
\toprule
Task & Total & Valid & Excluded & Valid ratio & Min & Median & P95 \\
\midrule
DiffReact & 100 & 100 & 0 & 1.0000 & 102.7 & 155.3 & 260.8 \\
Kolmogorov & 100 & 100 & 0 & 1.0000 & 9.059 & 37.39 & 150 \\
Burgers & 206 & 206 & 0 & 1.0000 & 1.253 & 3.041 & 7.492 \\
Advection & 206 & 206 & 0 & 1.0000 & 15.7 & 53.24 & 217.7 \\
\bottomrule
\end{tabular}
\caption{Sample-level response-mask statistics on the frozen test pairs. Directional response cosine excludes samples whose reference-response norm $\lVert r^\star\rVert_2$ is below the fixed threshold $\delta=10^{-8}$. Min, median, and P95 summarize the sample-level distribution of $\lVert r^\star\rVert_2$.}
\label{tab:response_mask_statistics}
\end{table*}

\section{Cross-System Results and Evidence Roles}

\subsection{Submitted Task-Wise Results}

Tables~\ref{tab:app_model_zoo_1}--\ref{tab:app_model_zoo_4} reproduce the
submitted cross-system comparison by task. These tables establish the primary
task--backbone observations used in the main paper. They show broad state and
rollout improvements while retaining the task- and architecture-dependent
response behavior emphasized in the main text. DiffReact exhibits the most
consistent response gains. Kolmogorov shows broader state and rollout
improvements but a mixed intrinsic-response pattern across backbones. Burgers
and Advection further show that response supervision is not uniformly
beneficial for every architecture and metric.

The task-wise tables should therefore be read as evidence for a
backbone-dependent training effect rather than as a claim of universal
intrinsic-response improvement. In particular, a better state or rollout
value does not imply that the same state predictor also has a better intrinsic
finite-step response.

\begin{figure*}[t]
\centering
\includegraphics[width=0.82\textwidth]{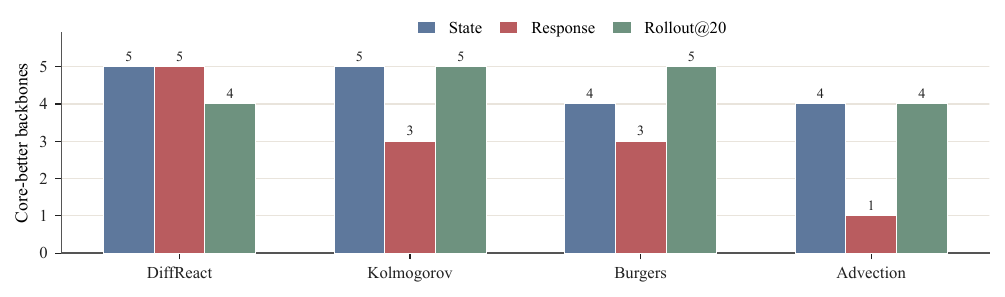}
\caption{Task-wise improvement coverage across the five submitted backbones. Bars count the task--backbone pairs for which TangentPDE-Core has the better point estimate than Aug-only for State, intrinsic response cosine, and Rollout@20. The figure summarizes the submitted task-wise tables and preserves the mixed response behavior; counts do not imply statistical significance.}
\label{fig:taskwise_improvement_coverage}
\end{figure*}

\begin{table}[tbp]
\centering
\small
\renewcommand{\arraystretch}{1.06}
\setlength{\tabcolsep}{2.2pt}
\begin{tabular*}{\columnwidth}{@{\extracolsep{\fill}}llrrr@{}}
\toprule
Model & Method & State $\downarrow$ & Resp. $\uparrow$ & Roll. $\downarrow$ \\
\midrule
CViT & Aug-only & 0.011589 & 0.1411 & 0.00690 \\
 & TangentPDE-Core & \textbf{0.009398} & \textbf{0.4721} & \textbf{0.00590} \\
FNO & Aug-only & 0.007529 & 0.4367 & 0.00413 \\
 & TangentPDE-Core & \textbf{0.005939} & \textbf{0.9358} & \textbf{0.00336} \\
LaMO & Aug-only & 0.007906 & 0.2455 & \textbf{0.00344} \\
 & TangentPDE-Core & \textbf{0.007188} & \textbf{0.4855} & 0.00348 \\
MambaNO & Aug-only & 0.012105 & 0.0722 & 0.00830 \\
 & TangentPDE-Core & \textbf{0.010433} & \textbf{0.9345} & \textbf{0.00740} \\
RNO & Aug-only & 0.008452 & 0.5431 & 0.00580 \\
 & TangentPDE-Core & \textbf{0.006316} & \textbf{0.9630} & \textbf{0.00440} \\
\bottomrule
\end{tabular*}
\caption{Expanded DiffReact results under the matched evaluation protocol. Bold marks the better value within each Aug-only/Core pair.}
\label{tab:app_model_zoo_1}
\end{table}

\begin{table}[tbp]
\centering
\small
\renewcommand{\arraystretch}{1.06}
\setlength{\tabcolsep}{2.2pt}
\begin{tabular*}{\columnwidth}{@{\extracolsep{\fill}}llrrr@{}}
\toprule
Model & Method & State $\downarrow$ & Resp. $\uparrow$ & Roll. $\downarrow$ \\
\midrule
CViT & Aug-only & 0.414042 & 0.3075 & 0.43090 \\
 & TangentPDE-Core & \textbf{0.353682} & \textbf{0.3192} & \textbf{0.26900} \\
FNO & Aug-only & 0.237184 & 0.7392 & 0.22747 \\
 & TangentPDE-Core & \textbf{0.237143} & \textbf{0.7798} & \textbf{0.22030} \\
LaMO & Aug-only & 0.213051 & \textbf{0.4187} & 0.38450 \\
 & TangentPDE-Core & \textbf{0.200136} & 0.4000 & \textbf{0.36410} \\
MambaNO & Aug-only & 0.488124 & 0.6189 & 0.80600 \\
 & TangentPDE-Core & \textbf{0.483679} & \textbf{0.6493} & \textbf{0.79490} \\
RNO & Aug-only & 0.223250 & \textbf{0.6769} & 0.39700 \\
 & TangentPDE-Core & \textbf{0.179350} & 0.5074 & \textbf{0.33470} \\
\bottomrule
\end{tabular*}
\caption{Expanded Kolmogorov results under the matched evaluation protocol. Bold marks the better value within each Aug-only/Core pair.}
\label{tab:app_model_zoo_2}
\end{table}

\begin{table}[tbp]
\centering
\small
\renewcommand{\arraystretch}{1.06}
\setlength{\tabcolsep}{2.2pt}
\begin{tabular*}{\columnwidth}{@{\extracolsep{\fill}}llrrr@{}}
\toprule
Model & Method & State $\downarrow$ & Resp. $\uparrow$ & Roll. $\downarrow$ \\
\midrule
CViT & Aug-only & 0.293270 & 0.2084 & 0.25000 \\
 & TangentPDE-Core & \textbf{0.265638} & \textbf{0.2191} & \textbf{0.24610} \\
FNO & Aug-only & \textbf{0.156602} & 0.3901 & 0.15502 \\
 & TangentPDE-Core & 0.157755 & \textbf{0.4533} & \textbf{0.15108} \\
LaMO & Aug-only & 0.287690 & \textbf{0.2064} & 0.25420 \\
 & TangentPDE-Core & \textbf{0.286906} & 0.2056 & \textbf{0.25270} \\
MambaNO & Aug-only & 0.683363 & \textbf{0.1263} & 0.72760 \\
 & TangentPDE-Core & \textbf{0.635693} & 0.1077 & \textbf{0.68070} \\
RNO & Aug-only & 0.321615 & 0.1989 & 0.28290 \\
 & TangentPDE-Core & \textbf{0.289794} & \textbf{0.2288} & \textbf{0.25390} \\
\bottomrule
\end{tabular*}
\caption{Expanded Burgers results under the matched evaluation protocol. Bold marks the better value within each Aug-only/Core pair.}
\label{tab:app_model_zoo_3}
\end{table}

\begin{table}[tbp]
\centering
\small
\renewcommand{\arraystretch}{1.06}
\setlength{\tabcolsep}{2.2pt}
\begin{tabular*}{\columnwidth}{@{\extracolsep{\fill}}llrrr@{}}
\toprule
Model & Method & State $\downarrow$ & Resp. $\uparrow$ & Roll. $\downarrow$ \\
\midrule
CViT & Aug-only & 0.546122 & \textbf{0.0471} & 0.06160 \\
 & TangentPDE-Core & \textbf{0.483750} & 0.0454 & \textbf{0.05960} \\
FNO & Aug-only & \textbf{0.129336} & \textbf{0.0899} & \textbf{0.03090} \\
 & TangentPDE-Core & 0.135071 & 0.0881 & 0.06828 \\
LaMO & Aug-only & 0.557364 & \textbf{0.0492} & 0.09400 \\
 & TangentPDE-Core & \textbf{0.555104} & 0.0475 & \textbf{0.07730} \\
MambaNO & Aug-only & 0.787117 & \textbf{0.0455} & 0.12610 \\
 & TangentPDE-Core & \textbf{0.775974} & 0.0403 & \textbf{0.09050} \\
RNO & Aug-only & 0.670729 & 0.0135 & 0.65600 \\
 & TangentPDE-Core & \textbf{0.636664} & \textbf{0.0184} & \textbf{0.60530} \\
\bottomrule
\end{tabular*}
\caption{Expanded Advection results under the matched evaluation protocol. Bold marks the better value within each Aug-only/Core pair.}
\label{tab:app_model_zoo_4}
\end{table}

\subsection{Supplementary Magnitude Diagnostics}

Tables~\ref{tab:canonical_diffreact}--\ref{tab:canonical_advection} retain the
submitted state, intrinsic finite-step response cosine, and Rollout@20 means,
and append an independent three-seed magnitude-error diagnostic. The
supplementary magnitude statistics describe only the added diagnostic and are
not uncertainty estimates for the submitted state, response-cosine, or rollout
values.

These tables therefore separate the identity of the main-paper quantities from
the identity of the added diagnostic column. Negative and mixed cells are
retained as task- and backbone-dependent evidence rather than converted into a
claim that every metric persists under one common schedule.

\begin{table*}[t]
\centering
\small
\renewcommand{\arraystretch}{1.06}
\setlength{\tabcolsep}{2.1pt}
\begin{tabular*}{\textwidth}{@{\extracolsep{\fill}}llrrrr@{}}
\toprule
Model & Method & State $\downarrow$ & Resp. $\uparrow$ & Supp. mag. $\downarrow$ & Roll. $\downarrow$ \\
\midrule
CViT & Aug-only & 0.011589 & 0.1411 & 0.3249 $\pm$ 0.0305 & 0.00690 \\
 & TangentPDE-Core & \textbf{0.009398} & \textbf{0.4721} & 0.3281 $\pm$ 0.0111 & \textbf{0.00590} \\
FNO & Aug-only & 0.007529 & 0.4367 & 0.8152 $\pm$ 0.0646 & 0.00413 \\
 & TangentPDE-Core & \textbf{0.005939} & \textbf{0.9358} & 0.3167 $\pm$ 0.3682 & \textbf{0.00336} \\
LaMO & Aug-only & 0.007906 & 0.2455 & 1.0623 $\pm$ 0.0634 & \textbf{0.00344} \\
 & TangentPDE-Core & \textbf{0.007188} & \textbf{0.4855} & 0.3807 $\pm$ 0.6121 & 0.00348 \\
MambaNO & Aug-only & 0.012105 & 0.0722 & 2.2330 $\pm$ 0.2587 & 0.00830 \\
 & TangentPDE-Core & \textbf{0.010433} & \textbf{0.9345} & 0.2033 $\pm$ 0.0044 & \textbf{0.00740} \\
RNO & Aug-only & 0.008452 & 0.5431 & 0.8445 $\pm$ 0.0443 & 0.00580 \\
 & TangentPDE-Core & \textbf{0.006316} & \textbf{0.9630} & 0.0112 $\pm$ 0.0039 & \textbf{0.00440} \\
\bottomrule
\end{tabular*}
\caption{DiffReact results. State, response cosine, and Rollout@20 reproduce the main-paper means; supplementary magnitude error is mean $\pm$ standard deviation from the independent three-seed diagnostic.}
\label{tab:canonical_diffreact}
\end{table*}

\begin{table*}[t]
\centering
\small
\renewcommand{\arraystretch}{1.06}
\setlength{\tabcolsep}{2.1pt}
\begin{tabular*}{\textwidth}{@{\extracolsep{\fill}}llrrrr@{}}
\toprule
Model & Method & State $\downarrow$ & Resp. $\uparrow$ & Supp. mag. $\downarrow$ & Roll. $\downarrow$ \\
\midrule
CViT & Aug-only & 0.414042 & 0.3075 & 1.7635 $\pm$ 0.0715 & 0.43090 \\
 & TangentPDE-Core & \textbf{0.353682} & \textbf{0.3192} & 0.6632 $\pm$ 0.3295 & \textbf{0.26900} \\
FNO & Aug-only & 0.237184 & 0.7392 & 1.0413 $\pm$ 0.0058 & 0.22747 \\
 & TangentPDE-Core & \textbf{0.237143} & \textbf{0.7798} & 1.0397 $\pm$ 0.0043 & \textbf{0.22030} \\
LaMO & Aug-only & 0.213051 & \textbf{0.4187} & 1.2073 $\pm$ 0.0397 & 0.38450 \\
 & TangentPDE-Core & \textbf{0.200136} & 0.4000 & 1.2221 $\pm$ 0.0345 & \textbf{0.36410} \\
MambaNO & Aug-only & 0.488124 & 0.6189 & 1.0302 $\pm$ 0.0055 & 0.80600 \\
 & TangentPDE-Core & \textbf{0.483679} & \textbf{0.6493} & 1.0881 $\pm$ 0.0437 & \textbf{0.79490} \\
RNO & Aug-only & 0.223250 & \textbf{0.6769} & 1.7261 $\pm$ 0.1585 & 0.39700 \\
 & TangentPDE-Core & \textbf{0.179350} & 0.5074 & 1.7674 $\pm$ 0.1716 & \textbf{0.33470} \\
\bottomrule
\end{tabular*}
\caption{Kolmogorov results. State, response cosine, and Rollout@20 reproduce the main-paper means; supplementary magnitude error is mean $\pm$ standard deviation from the independent three-seed diagnostic.}
\label{tab:canonical_kolmogorov}
\end{table*}

\begin{table*}[t]
\centering
\small
\renewcommand{\arraystretch}{1.06}
\setlength{\tabcolsep}{2.1pt}
\begin{tabular*}{\textwidth}{@{\extracolsep{\fill}}llrrrr@{}}
\toprule
Model & Method & State $\downarrow$ & Resp. $\uparrow$ & Supp. mag. $\downarrow$ & Roll. $\downarrow$ \\
\midrule
CViT & Aug-only & 0.293270 & 0.2084 & 0.5475 $\pm$ 0.0229 & 0.25000 \\
 & TangentPDE-Core & \textbf{0.265638} & \textbf{0.2191} & 0.3839 $\pm$ 0.0098 & \textbf{0.24610} \\
FNO & Aug-only & \textbf{0.156602} & 0.3901 & 0.4872 $\pm$ 0.0172 & 0.15502 \\
 & TangentPDE-Core & 0.157755 & \textbf{0.4533} & 0.2572 $\pm$ 0.0112 & \textbf{0.15108} \\
LaMO & Aug-only & 0.287690 & \textbf{0.2064} & 0.8402 $\pm$ 0.0204 & 0.25420 \\
 & TangentPDE-Core & \textbf{0.286906} & 0.2056 & 0.4180 $\pm$ 0.0041 & \textbf{0.25270} \\
MambaNO & Aug-only & 0.683363 & \textbf{0.1263} & 1.4627 $\pm$ 0.0515 & 0.72760 \\
 & TangentPDE-Core & \textbf{0.635693} & 0.1077 & 1.6136 $\pm$ 0.0398 & \textbf{0.68070} \\
RNO & Aug-only & 0.321615 & 0.1989 & 0.7605 $\pm$ 0.0566 & 0.28290 \\
 & TangentPDE-Core & \textbf{0.289794} & \textbf{0.2288} & 0.5202 $\pm$ 0.0242 & \textbf{0.25390} \\
\bottomrule
\end{tabular*}
\caption{Burgers results. State, response cosine, and Rollout@20 reproduce the main-paper means; supplementary magnitude error is mean $\pm$ standard deviation from the independent three-seed diagnostic.}
\label{tab:canonical_burgers}
\end{table*}

\begin{table*}[t]
\centering
\small
\renewcommand{\arraystretch}{1.06}
\setlength{\tabcolsep}{2.1pt}
\begin{tabular*}{\textwidth}{@{\extracolsep{\fill}}llrrrr@{}}
\toprule
Model & Method & State $\downarrow$ & Resp. $\uparrow$ & Supp. mag. $\downarrow$ & Roll. $\downarrow$ \\
\midrule
CViT & Aug-only & 0.546122 & \textbf{0.0471} & 2.4749 $\pm$ 0.0164 & 0.06160 \\
 & TangentPDE-Core & \textbf{0.483750} & 0.0454 & 2.2433 $\pm$ 0.0135 & \textbf{0.05960} \\
FNO & Aug-only & \textbf{0.129336} & \textbf{0.0899} & 2.5620 $\pm$ 0.0116 & \textbf{0.03090} \\
 & TangentPDE-Core & 0.135071 & 0.0881 & 2.5603 $\pm$ 0.0095 & 0.06828 \\
LaMO & Aug-only & 0.557364 & \textbf{0.0492} & 3.1964 $\pm$ 0.0715 & 0.09400 \\
 & TangentPDE-Core & \textbf{0.555104} & 0.0475 & 1.5833 $\pm$ 0.0871 & \textbf{0.07730} \\
MambaNO & Aug-only & 0.787117 & \textbf{0.0455} & 1.7060 $\pm$ 0.0369 & 0.12610 \\
 & TangentPDE-Core & \textbf{0.775974} & 0.0403 & 1.5748 $\pm$ 0.0245 & \textbf{0.09050} \\
RNO & Aug-only & 0.670729 & 0.0135 & 3.5990 $\pm$ 0.2384 & 0.65600 \\
 & TangentPDE-Core & \textbf{0.636664} & \textbf{0.0184} & 2.1184 $\pm$ 0.5829 & \textbf{0.60530} \\
\bottomrule
\end{tabular*}
\caption{Advection results. State, response cosine, and Rollout@20 reproduce the main-paper means; supplementary magnitude error is mean $\pm$ standard deviation from the independent three-seed diagnostic.}
\label{tab:canonical_advection}
\end{table*}

\begin{figure*}[t]
\centering
\includegraphics[width=0.94\textwidth]{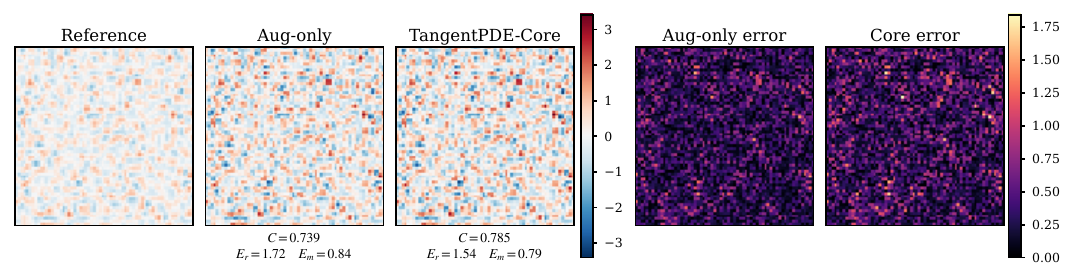}
\caption{Kolmogorov intrinsic-response fields for a representative canonical
FNO test query. Signed responses share one zero-centered color scale, and the
two absolute-error panels share a second scale. The query-selection rule is
median Core-minus-Aug-only response-cosine change over the frozen seed-1234
test set. The heatmaps show the time slice with the largest reference-response
norm. Annotated response cosine $C$, relative response error $E_r$, and
magnitude error $E_m$ are full-trajectory sample metrics computed using the
same definitions as the aggregate tables; they are not three-seed means.}
\label{fig:app_kolmogorov_response}
\end{figure*}

\begin{figure*}[t]
\centering
\includegraphics[width=0.94\textwidth]{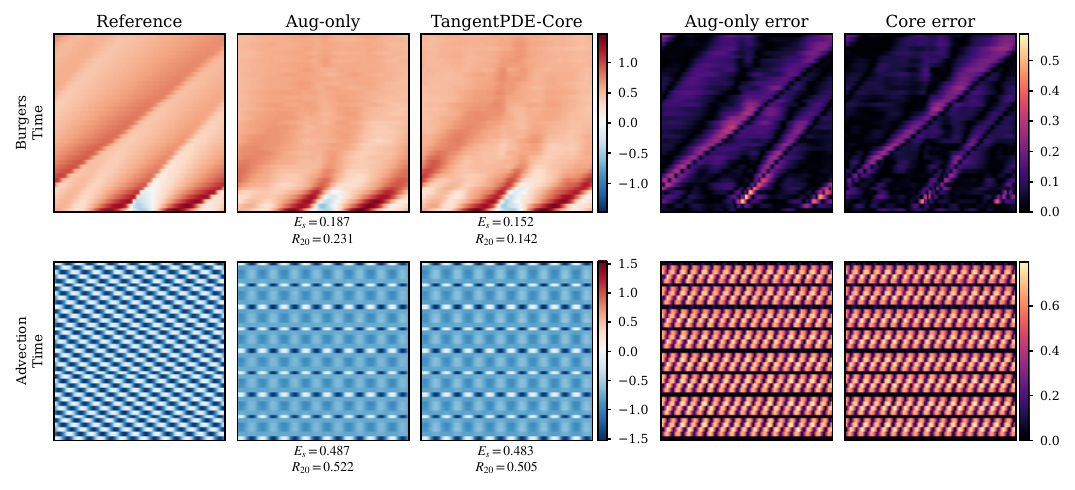}
\caption{Space--time state trajectories for representative Burgers and
Advection queries from the independent uniform-protocol FNO reruns. Rows use
task-specific shared state and absolute-error color scales. The annotations
report sample-level full-trajectory state relative $L_2$ error $E_s$ and
Rollout@20 error $R_{20}$ using the same definitions as the aggregate tables.
For each task, the displayed query has Core-minus-Aug-only state improvement
closest to the median over the frozen seed-1234 test set. These diagnostic
rerun fields do not alter the submitted task-wise point estimates.}
\label{fig:app_transport_state}
\end{figure*}

\subsection{Across-Setting Paired Analysis}

\begin{figure*}[t]
\centering
\includegraphics[width=0.86\textwidth]{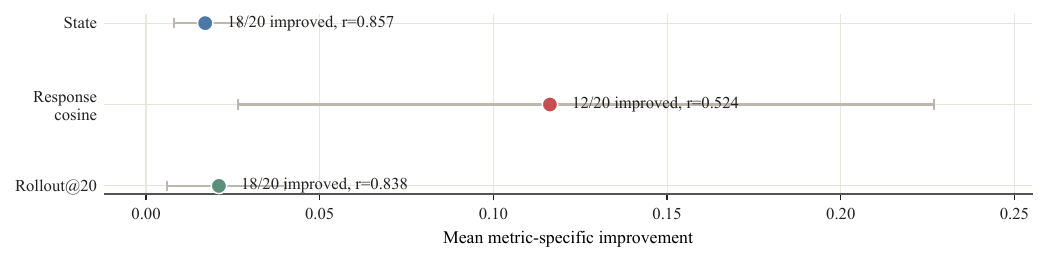}
\caption{Across-setting paired summary over the 20 submitted task--backbone settings. Points show mean metric-specific Core-over-Aug improvement; annotations report improved-setting counts. Positive values indicate improvement, and the figure is descriptive rather than a standalone inferential display.}
\label{fig:overall_paired_summary}
\end{figure*}

Figure~\ref{fig:overall_paired_summary} summarizes the 20 matched means
reported in the main paper. The comparison unit is one complete
task--backbone setting, preserving the paired Aug-only/Core comparison within
that setting. The reported changes use metric-specific improvement directions:
$\Delta_{\mathrm{state}}=E_{\mathrm{state}}(\mathrm{Aug})-
E_{\mathrm{state}}(\mathrm{Core})$,
$\Delta_{\mathrm{resp}}=C_{\mathrm{resp}}(\mathrm{Core})-
C_{\mathrm{resp}}(\mathrm{Aug})$, and
$\Delta_{\mathrm{roll}}=E_{\mathrm{roll}}(\mathrm{Aug})-
E_{\mathrm{roll}}(\mathrm{Core})$. Under these definitions, positive values
consistently denote improvement by TangentPDE-Core. The underlying paired
checks retain the same intervals and rank-based summaries as the source table,
while the figure foregrounds the descriptive cross-setting trend.

This analysis summarizes whether improvements are systematic across the
submitted settings. It does not treat grid points, time steps, or random seeds
as independent observations, and it does not convert a mixed per-backbone
response pattern into a universal claim.

\begin{table*}[t]
\centering
\small
\renewcommand{\arraystretch}{1.04}
\setlength{\tabcolsep}{3.0pt}
\caption{Paired analysis over the 20 submitted task--backbone settings. Positive changes denote Core improvement; intervals use 10,000 paired bootstrap resamples, and $p$ values use Holm-corrected one-sided Wilcoxon signed-rank tests.}
\label{tab:app_across_setting_paired}
\begin{tabular*}{\textwidth}{@{\extracolsep{\fill}}lrrrrrrrr@{}}
\toprule
Metric & Improved & \shortstack{Mean\\$\Delta$} &
\shortstack{Median\\$\Delta$} & \shortstack{Bootstrap\\95\% CI} &
$W$ & \shortstack{Raw\\$p$} & \shortstack{Holm\\$p$} &
$r_{\mathrm{rb}}$ \\
\midrule
State & 18/20 & 0.017041 & 0.003352 &
[0.007986, 0.026833] & 195.0 & 0.000131 & 0.000392 & 0.857 \\
Response cosine & 12/20 & 0.116285 & 0.011200 &
[0.026504, 0.226892] & 160.0 & 0.020000 & 0.020000 & 0.524 \\
Rollout@20 & 18/20 & 0.020988 & 0.005555 &
[0.006057, 0.039931] & 193.0 & 0.000197 & 0.000395 & 0.838 \\
\bottomrule
\end{tabular*}
\end{table*}

\section{Controlled Ablations and Competing Explanations}

Table~\ref{tab:app_budget} tests whether the observed gains can be explained
solely by seeing more endpoint states or by using a larger reference-query
budget. MSE-2N matches the number of supervised state observations, while
Two-point MSE uses the same paired endpoints and solver-query budget as Core
but omits response supervision. On DiffReact, Core improves all five metrics
beyond these controls. On Kolmogorov, the response and rollout gains are
accompanied by a small state-error trade-off.

\begin{table}[tbp]
\centering
\small
\renewcommand{\arraystretch}{1.06}
\setlength{\tabcolsep}{2.5pt}
\begin{tabular}{@{}lcccc@{}}
\toprule
Metric & MSE-1$\times$ & MSE-2N & Two-point & Core \\
\midrule
\multicolumn{5}{@{}l}{\textit{DiffReact}} \\
State $\downarrow$     & 0.009452 & 0.009445 & \second{0.009223} & \best{0.005908} \\
Resp. cos. $\uparrow$  & 0.2935 & 0.2944 & \second{0.2960} & \best{0.9308} \\
Resp. rel. $L_2$ $\downarrow$ & 0.9773 & 0.9768 & \second{0.9733} & \best{0.3758} \\
Mag. err. $\downarrow$ & 1.104 & 1.102 & \second{1.050} & \best{0.0997} \\
Roll.@20 $\downarrow$  & 0.004679 & 0.004687 & \second{0.004440} & \best{0.003573} \\
\midrule
\multicolumn{5}{@{}l}{\textit{Kolmogorov}} \\
State $\downarrow$     & 0.2810 & 0.2810 & 0.2810 & 0.2846 \\
Resp. cos. $\uparrow$  & 0.7204 & 0.7204 & \second{0.7205} & \best{0.7528} \\
Resp. rel. $L_2$ $\downarrow$ & 4.053 & \second{4.052} & \second{4.052} & \best{3.918} \\
Mag. err. $\downarrow$ & 1.108 & 1.108 & 1.108 & \best{1.095} \\
Roll.@20 $\downarrow$  & 0.2960 & 0.2960 & \second{0.2959} & \best{0.2945} \\
\bottomrule
\end{tabular}
\caption{Budget-matched FNO controls. MSE-2N matches the state count; Two-point uses the same paired endpoints and reference-query budget as Core but omits response supervision.}
\label{tab:app_budget}
\end{table}

Table~\ref{tab:app_mechanism} separates direct response-vector regression from
the direction--magnitude decomposition. Direct Response MSE applies one
Euclidean objective to the undecomposed finite-step response vector. The
objective rows therefore test whether the decomposition itself matters, while
the pair rows test whether exact endpoint correspondence contributes beyond
observing the same endpoint states.

The pair-shuffled control is interpreted narrowly. It preserves the available
endpoint states, pair count, and optimization budget while disrupting the
declared relational correspondence. Consequently, it tests whether correct
pair structure contributes beyond a response-shaped regularization effect; it
does not establish that every mismatch is uninformative or that the same
mechanism must hold for every task and backbone.

\begin{table}[tbp]
\centering
\small
\renewcommand{\arraystretch}{1.04}
\setlength{\tabcolsep}{2.4pt}
\begin{tabular}{@{}lccc@{}}
\toprule
Setting & State $\downarrow$ & Resp. cos. $\uparrow$ & Roll.@20 $\downarrow$ \\
\midrule
\multicolumn{4}{@{}l}{\textit{Objective composition: DiffReact}} \\
State only   & 0.008536 & 0.5360 & 0.0059 \\
State + Direct Resp. MSE & 0.008095 & 0.5953 & \second{0.0050} \\
State + Dir. & \second{0.007107} & \best{0.9640} & \best{0.0044} \\
State + Mag. & 0.008715 & 0.4311 & 0.0065 \\
Full Core    & \best{0.006316} & \second{0.9630} & \best{0.0044} \\
\midrule
\multicolumn{4}{@{}l}{\textit{Objective composition: Kolmogorov}} \\
State only   & 0.223495 & \best{0.6775} & 0.3959 \\
State + Direct Resp. MSE & \best{0.179553} & 0.5073 & \best{0.3350} \\
State + Dir. & \second{0.179643} & 0.5089 & \second{0.3351} \\
State + Mag. & 0.179648 & 0.5077 & \second{0.3351} \\
Full Core    & 0.179683 & \second{0.5090} & \second{0.3351} \\
\midrule
\multicolumn{4}{@{}l}{\textit{Pair correspondence: DiffReact}} \\
State baseline & 0.008267 & 0.5349 & 0.0057 \\
Pair-shuffled  & \second{0.006797} & \second{0.9297} & \second{0.0050} \\
True Core      & \best{0.006316} & \best{0.9630} & \best{0.0044} \\
\midrule
\multicolumn{4}{@{}l}{\textit{Pair correspondence: Kolmogorov}} \\
State baseline & 0.222942 & \best{0.6783} & 0.3936 \\
Pair-shuffled  & \second{0.179625} & 0.5061 & \best{0.3350} \\
True Core      & \best{0.179574} & \second{0.5068} & \best{0.3350} \\
\bottomrule
\end{tabular}
\caption{RNO mechanism controls. Objective rows isolate direct response regression, direction, and magnitude terms; pair rows break or retain endpoint correspondence.}
\label{tab:app_mechanism}
\end{table}

\begin{figure*}[t]
\centering
\includegraphics[width=0.86\textwidth]{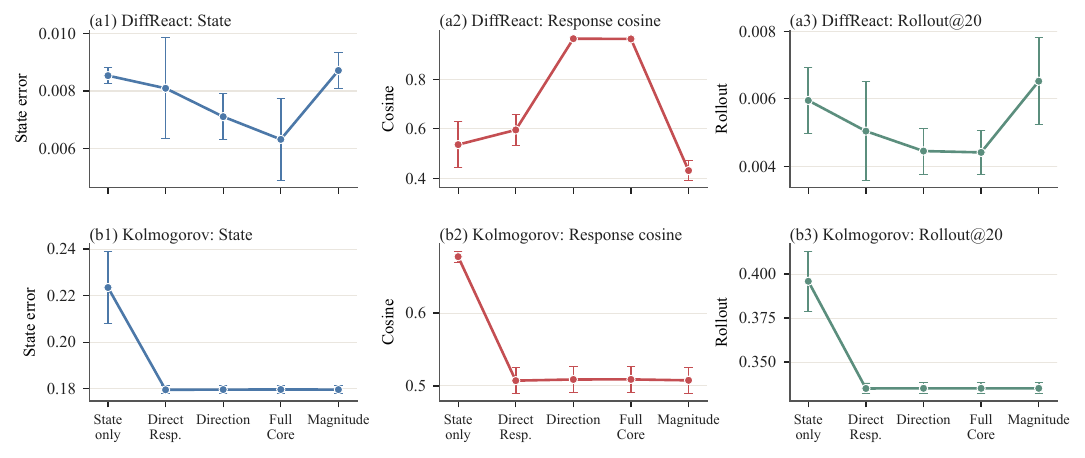}
\caption{Objective-composition study on DiffReact and Kolmogorov. Points show three-seed means with sample-standard-deviation error bars for State, intrinsic response cosine, and Rollout@20.}
\label{fig:objective_composition}
\end{figure*}

Figure~\ref{fig:objective_composition} visualizes the seed-level objective
composition rows from the same RNO mechanism study. The plot emphasizes the
relative behavior of the objective variants on DiffReact and Kolmogorov, while
the source values are kept unchanged.

\section{Layer-Selective Tangent Head: Semantics and Diagnostics}

Table~\ref{tab:app_lsth} distinguishes the intrinsic response of the shared
state predictor from the auxiliary response estimate produced by LSTH. Let
$G_{\theta'}$ denote the shared state predictor after LSTH continuation. The
frozen main-text semantics are
\begin{align}
 r_{\mathrm{op}}
 &= \frac{G_{\theta'}(z^+)-G_{\theta'}(z)}{\epsilon},\\
 \Delta h_k &= \frac{h_k(z^+)-h_k(z)}{\epsilon},\\
 c_\phi &= \gamma\sum_{k\in\mathcal K}A_k(\Delta h_k),\\
 r_{\mathrm{aux}} &= r_{\mathrm{op}}+c_\phi.
\end{align}
The correction is not added to either state output. The auxiliary metric is
defined only for LSTH and is therefore not compared with an implicitly
substituted intrinsic metric. Standard state prediction and intrinsic-response
inference bypass the adapters; the selected hidden-difference adapters execute
only when the auxiliary response estimate is requested.

LSTH reuses the Core response losses and regularizes the auxiliary correction
through
\begin{align}
 \mathcal L_{\mathrm{corr}}
 &= \frac{1}{B}\sum_{i=1}^{B}
 \left(\frac{\lVert c_{\phi,i}\rVert_2}
 {\lVert r_{\mathrm{op},i}\rVert_2+\delta}\right)^2,\\
 \mathcal L_{\mathrm{adapter}}
 &= 0.1\,\mathcal L_{\mathrm{dir}}(r_{\mathrm{aux}},r^\star)\notag\\
 &\quad +0.01\,\mathcal L_{\mathrm{mag}}(r_{\mathrm{aux}},r^\star)
 +10^{-4}\mathcal L_{\mathrm{corr}}.
\end{align}
The correction regularizer limits the auxiliary residual magnitude relative to
the intrinsic response during LSTH continuation. It is part of the LSTH
training path, whereas the over-correction value reported later is a separate
supplementary diagnostic.

To diagnose whether the optional correction dominates the intrinsic response,
we additionally report the implementation-defined over-correction statistic
\begin{equation}
 E_{\mathrm{over}}=
 \frac{1}{N_{\mathrm{valid}}}\sum_{i:\,\lVert r_i^\star\rVert_2\ge\delta}
 \frac{\lVert c_{\phi,i}\rVert_2}{\lVert r_{\mathrm{op},i}\rVert_2+\delta}.
\end{equation}
This sample-level quantity is averaged over valid response samples, uses the
same reference-response mask as the directional response metric, and is
diagnostic only. It is not used for training, checkpoint selection, or the
main-paper comparisons. Larger values indicate a larger auxiliary correction
relative to the intrinsic state-output response.

LSTH continuation uses a 20\% adapter-only warmup. During the remaining
updates, the selected shared blocks and output head receive auxiliary-gradient
weights $\mu_{\mathrm{block}}=0.05$ and $\mu_{\mathrm{head}}=0.02$, with
learning-rate multipliers 0.02 and 0.05, respectively. The two shared-parameter
gradient coefficients ramp linearly from zero during the first 20\% of the
second stage. The adapter learning-rate multiplier is 1.0, and the residual
scale is fixed at $\gamma=0.05$.

The LSTH study uses separate task-specific compact FNO configurations and is
not parameterized by the width-32, three-block FNO reported for the breadth
study in Table~\ref{tab:app_backbone_architecture}. Both LSTH configurations
use two FNO layers with hidden width 8, and the two available hidden stages are
used by the auxiliary head. Table~\ref{tab:app_lsth_overhead} reports the
parameter and optional auxiliary-response evaluation overhead for these two
compact FNO configurations. The latency baseline is the intrinsic
paired-response path of the same trained LSTH model, not state-only inference
and not the intrinsic path of a separate static model.

\begin{table*}[t]
\centering
\small
\renewcommand{\arraystretch}{1.08}
\setlength{\tabcolsep}{3.0pt}
\caption{LSTH parameter and auxiliary-response evaluation overhead. Base
parameters refer to the shared FNO state predictor. Added parameters count
only newly introduced hidden-difference adapters and correction heads; shared
backbone blocks and the output projection that participate in continuation
training are not counted as added parameters. Latency compares intrinsic
paired-response evaluation with auxiliary-response evaluation for the same
LSTH model. Measurements use batch size 4 on one NVIDIA GeForce RTX 4080
SUPER, with 50 warmup iterations, 200 measured iterations, and five repeats.
The displayed latencies are preserved aggregate means.}
\label{tab:app_lsth_overhead}
\begin{tabular*}{\textwidth}{@{\extracolsep{\fill}}lrrrrrrr@{}}
\toprule
Task & \shortstack{Base\\params} & \shortstack{Added\\params} & \shortstack{Total\\params} & \shortstack{Param.\\overhead} &
\shortstack{Intrinsic\\(ms/batch)} & \shortstack{Auxiliary\\(ms/batch)} & \shortstack{Latency\\overhead} \\
\midrule
DiffReact   & 7,526 & 1,996 & 9,522 & 26.52\% & 3.7778 & 4.2479 & 12.44\% \\
Kolmogorov  & 6,133 &   538 & 6,671 &  8.77\% & 3.6005 & 4.1841 & 16.21\% \\
\bottomrule
\end{tabular*}
\end{table*}

The parameter count uses unique \texttt{nn.Parameter} objects, includes biases,
and excludes buffers, optimizer states, activations, and checkpoint metadata.
The latency timing excludes data loading and host-to-device transfer. Raw
per-repeat timing vectors and explicit precision metadata were not retained,
so only aggregate mean latency is reported; median and P95 are not reported.
These measurements characterize only the optional auxiliary paired-response
path and do not change the cost of standard state-only inference.

The comparison among Static, State FT, Weak FT, and LSTH separates two
questions. State and rollout metrics evaluate the shared state predictor,
whereas the auxiliary response metric evaluates the dedicated LSTH execution
path. An auxiliary-response gain must therefore not be interpreted as evidence
that LSTH improves the intrinsic finite-step response of the same shared state
operator.

\begin{table}[tbp]
\centering
\small
\renewcommand{\arraystretch}{1.08}
\setlength{\tabcolsep}{1.8pt}
\begin{tabular}{@{}lcccc@{}}
\toprule
Variant & State $\downarrow$ & Intr. resp. $\uparrow$ & Aux. resp. $\uparrow$ & Roll. $\downarrow$ \\
\midrule
\multicolumn{5}{@{}l}{\textit{DiffReact}} \\
Static   & 0.0059 & \second{0.9433} & \na & 0.0035 \\
State FT & \best{0.0032} & 0.9410 & \na & \best{0.0018} \\
Weak FT  & 0.0037 & \best{0.9440} & \na & 0.0021 \\
LSTH     & \second{0.0036} & 0.9431 & 0.9779 & \second{0.0019} \\
\midrule
\multicolumn{5}{@{}l}{\textit{Kolmogorov}} \\
Static   & 0.2809 & \best{0.7572} & \na & 0.2941 \\
State FT & \best{0.2749} & 0.7281 & \na & 0.2887 \\
Weak FT  & \second{0.2795} & \second{0.7525} & \na & \second{0.2886} \\
LSTH     & \best{0.2749} & 0.7281 & 0.9071 & \best{0.2885} \\
\bottomrule
\end{tabular}
\caption{Layer-selective response-aware fine-tuning. Intrinsic response is obtained from the shared state predictor, while auxiliary response is defined only for LSTH.}
\label{tab:app_lsth}
\end{table}

\subsection{Supplementary LSTH Over-Correction Diagnostic}

Table~\ref{tab:canonical_lsth_summary} reproduces the submitted state,
intrinsic-response, auxiliary-response, and rollout means, and appends an
independent three-seed over-correction diagnostic for LSTH. The added
diagnostic does not reconstruct uncertainty around the submitted main metrics.

Over-correction is defined only for LSTH because the other variants do not
execute the auxiliary correction path. Dashes therefore denote undefined or
inapplicable quantities rather than numerical zero. The reported standard
deviation and confidence interval apply only to the supplementary
over-correction diagnostic column.

\begin{table*}[t]
\centering
\small
\renewcommand{\arraystretch}{1.08}
\setlength{\tabcolsep}{1.8pt}
\begin{tabular}{@{}lcccccc@{}}
\toprule
Variant & State $\downarrow$ & Intr. resp. $\uparrow$ & Aux. resp. $\uparrow$ & Roll. $\downarrow$ & Supp. over-corr. $\downarrow$ & 95\% CI \\
\midrule
\multicolumn{7}{@{}l}{\textit{DiffReact}} \\
Static   & 0.0059 & \second{0.9433} & \na & 0.0035 & \na & \na \\
State FT & \best{0.0032} & 0.9410 & \na & \best{0.0018} & \na & \na \\
Weak FT  & 0.0037 & \best{0.9440} & \na & 0.0021 & \na & \na \\
LSTH     & \second{0.0036} & 0.9431 & 0.9779 & \second{0.0019} & 0.0000 $\pm$ 0.0000 & [0.0000, 0.0000] \\
\midrule
\multicolumn{7}{@{}l}{\textit{Kolmogorov}} \\
Static   & 0.2809 & \best{0.7572} & \na & 0.2941 & \na & \na \\
State FT & \best{0.2749} & 0.7281 & \na & 0.2887 & \na & \na \\
Weak FT  & \second{0.2795} & \second{0.7525} & \na & \second{0.2886} & \na & \na \\
LSTH     & \best{0.2749} & 0.7281 & 0.9071 & \best{0.2885} & 0.3300 $\pm$ 0.0100 & [0.3052, 0.3548] \\
\bottomrule
\end{tabular}
\caption{LSTH supplementary over-correction diagnostic. State, intrinsic response, auxiliary response, and rollout reproduce the submitted means. Supplementary over-correction is an independent three-seed diagnostic and is defined only for LSTH; dashes denote undefined or inapplicable quantities, and its uncertainty is a two-sided 95\% Student-$t$ interval over three seeds.}
\label{tab:canonical_lsth_summary}
\end{table*}

\section{Computational Resources and Query Accounting}

\begin{table}[tbp]
\centering
\small
\setlength{\tabcolsep}{2pt}
\begin{tabular*}{\columnwidth}{@{\extracolsep{\fill}}lrrrr@{}}
\toprule
Task & Pairs & $\rho_{\mathrm{tr}}$ & Overall & Calls \\
\midrule
DiffReact & 200/50/100 & 100\% & 100\% & 350 \\
Kolmogorov & 200/50/100 & 100\% & 100\% & 350 \\
Burgers & 1638/204/206 & 16.38\% & 20.48\% & 2,048 \\
Advection & 1638/204/206 & 16.38\% & 20.48\% & 2,048 \\
\bottomrule
\end{tabular*}
\caption{Perturbation and reference-query accounting. $\rho_{\mathrm{tr}}$ is the fraction of training anchors receiving paired supervision; Overall paired is the fraction over all available anchors.}
\label{tab:app_protocol}
\end{table}

\IfFileExists{tables/table_unseen_cost.tex}{\begin{table}[tbp]
\centering
\small
\renewcommand{\arraystretch}{1.08}
\setlength{\tabcolsep}{1.2pt}
\begin{tabular*}{\columnwidth}{@{\extracolsep{\fill}}lccccc@{}}
\toprule
Task & Time (min) & \shortstack{95\% CI\\hw} & Peak (GiB) & \shortstack{95\% CI\\hw} & GPUs \\
\midrule
DiffReact & 6.4 $\pm$ 2.1 & [4.8, 8.0] & 0.18 $\pm$ 0.00 & [0.18, 0.18] & 1 \\
Kolmogorov & 7.9 $\pm$ 2.2 & [6.2, 9.6] & 0.09 $\pm$ 0.00 & [0.09, 0.09] & 1 \\
\bottomrule
\end{tabular*}
\caption{Measured cost of the unseen-direction runs. Time and peak allocated GPU memory are mean $\pm$ sample standard deviation per run, with two-sided 95\% Student-$t$ intervals over the nine objective--seed runs for each task.}
\label{tab:app_unseen_cost}
\end{table}

}{}

These intervals are descriptive summaries over nine heterogeneous objective--seed runs and should not be interpreted as uncertainty intervals for one fixed objective. Table~\ref{tab:pair_runtime} reports measured response-pair generation cost.
Each accepted pair adds one perturbed-endpoint solver call because its anchor
label is reused. DiffReact and Kolmogorov are regenerated from their frozen
perturbed inputs and numerically checked against the stored endpoints;
Burgers and Advection use the original complete source-generation logs.

\begin{table}[tbp]
\centering
\small
\setlength{\tabcolsep}{2.5pt}
\begin{tabular}{@{}lrrrl@{}}
\toprule
Task & Source & Add. calls & Time (s) & Execution \\
\midrule
Advection & 10000 & 10000 & 10.0 & 1 GPU \\
Burgers & 10000 & 10000 & 24.0 & 1 GPU \\
DiffReact & 350 & 350 & 25.1 & 4 CPU workers \\
Kolmogorov & 350 & 350 & 6.8 & 4 CPU workers \\
\bottomrule
\end{tabular}
\caption{Measured response-pair generation cost. DiffReact and Kolmogorov regenerate the 350 frozen perturbed endpoints. Burgers and Advection report the original 10,000-pair source generation before selecting the submitted subset. Each entry is one wall-clock measurement including output writing.}
\label{tab:pair_runtime}
\end{table}

Except for the separate LSTH latency microbenchmark in
Table~\ref{tab:app_lsth_overhead}, experiments were executed on a Linux 6.8
system with two Intel Xeon Gold 6530 processors, approximately 512~GiB of
system memory, and NVIDIA GeForce RTX 5090 GPUs with 32~GiB memory per device. The software environment used Python
3.10.12, PyTorch 2.12.0 with CUDA 13.0, NVIDIA driver 580.126.20, NumPy
2.2.6, and h5py 3.16.0. Independent single-GPU runs may be executed
concurrently; the number of GPUs visible to the host is not treated as the
number used by one training run.

The reported costs refer to different stages and should not be conflated.
Pair-generation time measures reference-solver data construction, whereas the
unseen-direction cost measures execution of a diagnostic evaluation workflow.
Neither quantity is a proxy for state-predictor training time or model
inference latency.

\subsection{Additional Burgers Objective Analysis}

\begin{figure}[tbp]
\centering
\includegraphics[width=\columnwidth]{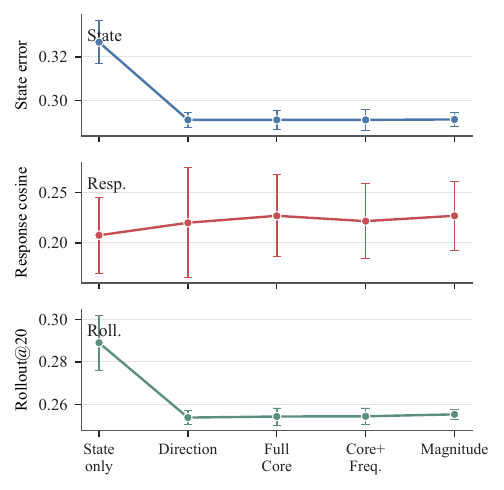}
\caption{Additional RNO objective ablation on Burgers. Points show three-seed means for State, response cosine, and Rollout@20 across the objective variants; error bars denote sample standard deviation. The plot is intended to compare objective trends rather than to foreground inferential statistics.}
\label{fig:burgers_objective_ablation}
\end{figure}

Figure~\ref{fig:burgers_objective_ablation} presents the Burgers objective
study as an
additional matched RNO ablation. It is separate from the cross-architecture
comparison and tests whether adding a frequency-response term changes the
direction--magnitude objective on nonlinear transport.

The frequency-response row is interpreted as a local objective comparison
within this Burgers configuration. It does not establish that frequency
supervision is uniformly beneficial across the other tasks or backbones.

\section{Robustness to Perturbation Changes}

\IfFileExists{tables/table_unseen_direction.tex}{%
The unseen-direction study separates perturbation families between training
and testing while preserving the original anchor splits, optimization budget,
seeds, and validation-state checkpoint criterion.
Table~\ref{tab:app_unseen_direction} therefore measures transfer across the
tested perturbation families rather than in-distribution response fitting.
It does not evaluate arbitrary directions outside the declared construction.
\begin{table*}[t]
\centering
\small
\renewcommand{\arraystretch}{1.08}
\setlength{\tabcolsep}{2.0pt}
\begin{tabular}{@{}llccccc@{}}
\toprule
Task & Method & State $\downarrow$ & Resp. cos. $\uparrow$ & Rel. resp. $\downarrow$ & Mag. err. $\downarrow$ & Roll.@20 $\downarrow$ \\
\midrule
DiffReact & Aug-only & 0.010514 $\pm$ 0.000405 & 0.1667 $\pm$ 0.0784 & 1.0227 $\pm$ 0.0643 & 1.5896 $\pm$ 0.6075 & 0.00580 $\pm$ 0.00133 \\
DiffReact & Two-point MSE & 0.010411 $\pm$ 0.000500 & 0.1707 $\pm$ 0.0680 & 1.0208 $\pm$ 0.0652 & 1.5491 $\pm$ 0.5881 & 0.00564 $\pm$ 0.00146 \\
DiffReact & TangentPDE-Core & \best{0.008211} $\pm$ 0.000776 & \best{0.8834} $\pm$ 0.0034 & \best{0.6393} $\pm$ 0.0108 & \best{0.2504} $\pm$ 0.0095 & \best{0.00491} $\pm$ 0.00034 \\
Kolmogorov & Aug-only & 0.286750 $\pm$ 0.003532 & 0.5059 $\pm$ 0.0021 & 9.8993 $\pm$ 0.0603 & 2.3386 $\pm$ 0.0056 & 0.31424 $\pm$ 0.01693 \\
Kolmogorov & Two-point MSE & 0.286668 $\pm$ 0.003640 & 0.5059 $\pm$ 0.0021 & 9.9005 $\pm$ 0.0588 & 2.3387 $\pm$ 0.0055 & 0.31396 $\pm$ 0.01725 \\
Kolmogorov & TangentPDE-Core & \best{0.286205} $\pm$ 0.005883 & \best{0.5419} $\pm$ 0.0016 & \best{9.7912} $\pm$ 0.0321 & \best{2.3318} $\pm$ 0.0031 & \best{0.30824} $\pm$ 0.02328 \\
\bottomrule
\end{tabular}
\par\vspace{15pt}
\begin{tabular}{@{}llccccc@{}}
\toprule
Task & Method & State CI & Resp. CI & Rel. resp. CI & Mag. CI & Roll. CI \\
\midrule
DiffReact & Aug-only & 0.001007 & 0.1948 & 0.1596 & 1.5091 & 0.00331 \\
DiffReact & Two-point MSE & 0.001243 & 0.1690 & 0.1620 & 1.4610 & 0.00363 \\
DiffReact & TangentPDE-Core & 0.001926 & 0.0086 & 0.0269 & 0.0235 & 0.00085 \\
Kolmogorov & Aug-only & 0.008773 & 0.0052 & 0.1498 & 0.0140 & 0.04205 \\
Kolmogorov & Two-point MSE & 0.009042 & 0.0051 & 0.1460 & 0.0137 & 0.04284 \\
Kolmogorov & TangentPDE-Core & 0.014615 & 0.0039 & 0.0797 & 0.0078 & 0.05783 \\
\bottomrule
\end{tabular}
\caption{FNO evaluation on perturbation families excluded from training. The upper panel reports mean $\pm$ sample standard deviation over three seeds; the lower panel gives the half-width of the corresponding two-sided 95\% Student-$t$ interval. Bold marks the best mean per task.}
\label{tab:app_unseen_direction}
\end{table*}

On DiffReact, Core improves all five reported metrics and raises response
cosine from 0.1667 for Aug-only and 0.1707 for Two-point MSE to 0.8834. On
Kolmogorov, Core provides a smaller response-cosine gain, from approximately
0.506 to 0.542; the state and rollout differences are small relative to their
across-seed variation. The transfer benefit is therefore task dependent.
}{}

\begin{figure}[!htbp]
\centering
\includegraphics[width=0.92\linewidth]{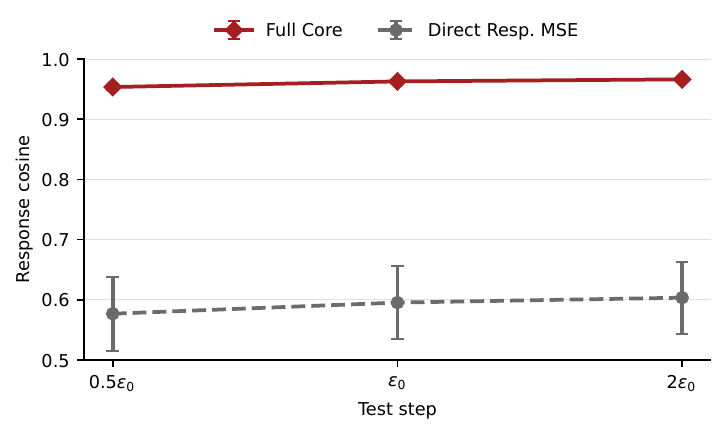}
\caption{Same-source DiffReact altered-step response evaluation using fixed
checkpoints. Points show the three-seed mean and error bars show the sample
standard deviation. The comparison evaluates transfer across the three tested
step sizes within the same perturbation family.}
\label{fig:app_step}
\end{figure}

The altered-step experiment holds the DiffReact anchors, perturbation
directions, normalization, and checkpoints fixed. The reaction--diffusion
solver regenerates each endpoint at $0.5\epsilon$, $\epsilon$, and
$2\epsilon$. Table~\ref{tab:app_epsilon} and
Figure~\ref{fig:app_step} show that the decomposed objective retains high
response alignment across all three test steps, whereas Direct Response MSE
remains substantially less aligned. Because the perturbation family and
checkpoints are fixed, this experiment tests same-source step-size transfer
only; it does not establish step-size invariance for the other PDE systems.

\begin{table}[!hbp]
\centering
\small
\renewcommand{\arraystretch}{1.08}
\setlength{\tabcolsep}{2.0pt}
\begin{tabular}{@{}llrr@{}}
\toprule
Objective & Test step & Resp. cos. $\uparrow$ & Mag. err. $\downarrow$ \\
\midrule
Full Core & $0.5\epsilon$ & 0.9537 $\pm$ 0.0041 & 0.0220 $\pm$ 0.0051 \\
Full Core & $\epsilon$ & 0.9630 $\pm$ 0.0035 & 0.0146 $\pm$ 0.0051 \\
Full Core & $2\epsilon$ & 0.9665 $\pm$ 0.0036 & 0.0191 $\pm$ 0.0056 \\
Direct Resp. MSE & $0.5\epsilon$ & 0.5768 $\pm$ 0.0616 & 0.8388 $\pm$ 0.0452 \\
Direct Resp. MSE & $\epsilon$ & 0.5953 $\pm$ 0.0610 & 0.8716 $\pm$ 0.0735 \\
Direct Resp. MSE & $2\epsilon$ & 0.6034 $\pm$ 0.0599 & 0.8858 $\pm$ 0.0878 \\
\bottomrule
\end{tabular}
\par\vspace{15pt}
\begin{tabular}{@{}llrr@{}}
\toprule
Objective & Test step & \shortstack{Resp. 95\%\\CI hw} & \shortstack{Mag. 95\%\\CI hw} \\
\midrule
Full Core & $0.5\epsilon$ & 0.0102 & 0.0127 \\
Full Core & $\epsilon$ & 0.0086 & 0.0127 \\
Full Core & $2\epsilon$ & 0.0088 & 0.0140 \\
Direct Resp. MSE & $0.5\epsilon$ & 0.1531 & 0.1123 \\
Direct Resp. MSE & $\epsilon$ & 0.1514 & 0.1826 \\
Direct Resp. MSE & $2\epsilon$ & 0.1489 & 0.2181 \\
\bottomrule
\end{tabular}
\caption{DiffReact same-source altered-step evaluation. Fixed checkpoints are evaluated on solver-regenerated endpoints at each test step. The upper panel reports mean $\pm$ sample standard deviation over three seeds; the lower panel gives the corresponding two-sided 95\% Student-$t$ CI half-width.}
\label{tab:app_epsilon}
\end{table}

\section{Scope and Additional Limitations}

TangentPDE learns finite-step secants along sampled perturbation families, not
a complete Jacobian or a uniformly accurate local linearization in every
direction. Improvements in state, intrinsic response, and rollout are related
but distinct outcomes; improvement in one metric does not logically imply
improvement in the others. The submitted results and the independent reruns
therefore retain their negative and mixed cells rather than collapsing them
into one universal claim.

Response gains remain task and backbone dependent. The unseen-direction study
covers only the declared held-out perturbation families, and the altered-step
evidence is limited to the evaluated same-source DiffReact construction.
Linear rescaling of an existing Kolmogorov pair is not a new PDE solve and is
excluded from the altered-step evidence. These boundaries are stated to keep
the appendix aligned with the implemented diagnostics rather than to broaden
the empirical claim.

LSTH provides a dedicated auxiliary response estimate rather than a universal
intrinsic-response improvement. Its auxiliary correction must be validated per
task because a residual that is useful for one system can over-correct another.
The LSTH auxiliary metric is consequently reported only on the execution path
for which it is defined. In particular, the auxiliary head is treated as a
separate diagnostic output, not as a direct correction to the reported state
operator. This distinction is kept explicit because the appendix contains both
shared-state metrics and auxiliary-response quantities.

Finally, seed-level summaries are based on three training seeds and should be
read as descriptive optimization variability rather than high-powered
single-configuration significance tests. Pair-generation time, diagnostic-run
time, training time, and inference latency measure different computational
stages and are not interchangeable. The resource tables therefore report
accounting for the stage named in each caption: query construction, diagnostic
evaluation, and auxiliary-head inference are intentionally separated.

The visual summaries preserve the metric directions and mixed outcomes of the
underlying submitted results. Exact protocol and accounting quantities remain
reported in tables. These conventions make the appendix a reproducibility and
audit companion to the main paper rather than an additional source of broader
claims.

The ordering of the appendix follows the evidence type rather than a single
performance ranking. Configuration and budget tables establish comparability,
mechanism tables separate objective components, resource tables document the
query and execution accounting, and the final robustness tables record
perturbation-family and step-size checks. This order is intended to make clear
which quantities are directly comparable and which are protocol-specific.

\FloatBarrier
\clearpage
\begin{figure*}[p]
\centering
\includegraphics[width=0.94\textwidth]{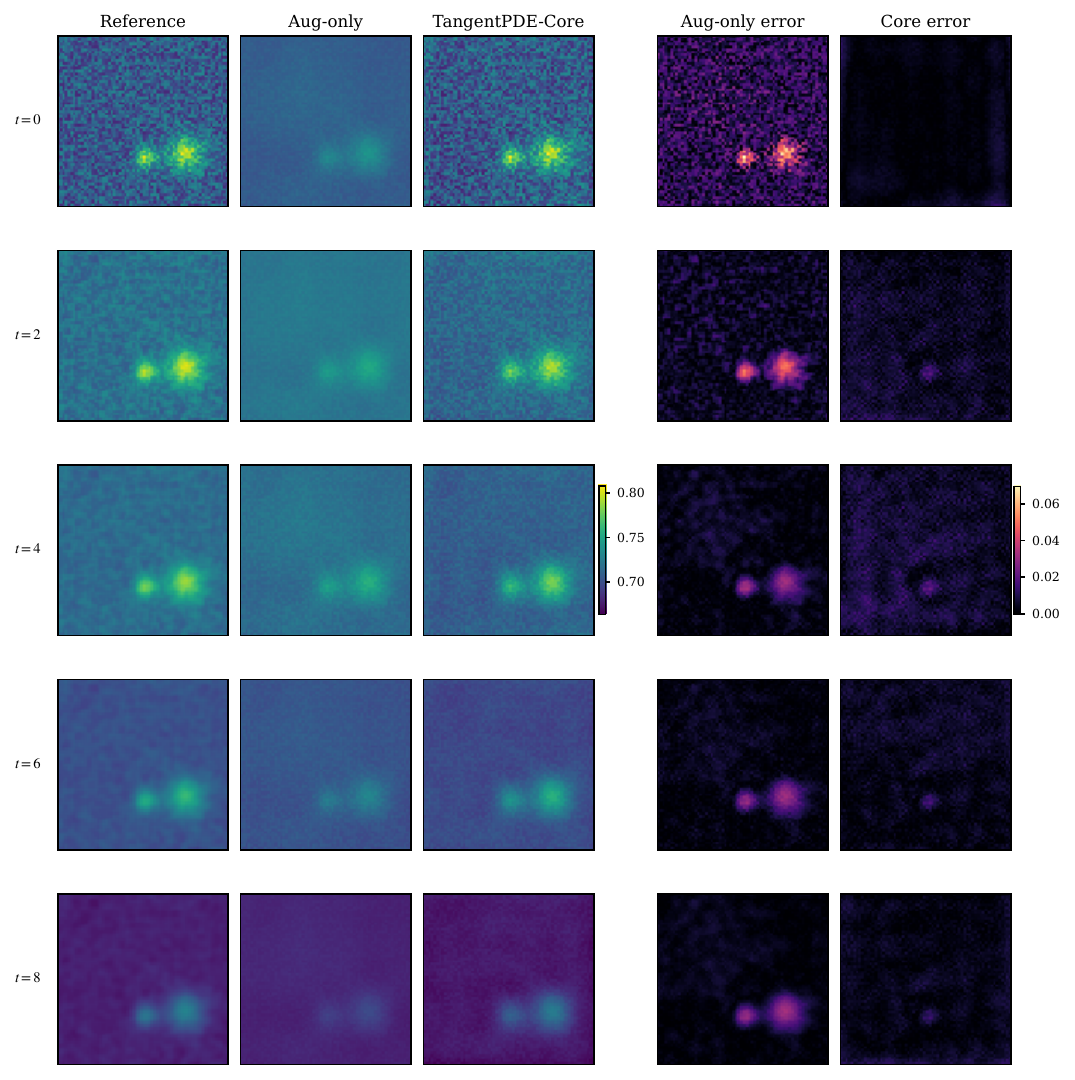}
\caption{Early-time DiffReact state fields for the first output channel. The
columns show the reference field, Aug-only and TangentPDE-Core predictions,
and their absolute errors at five fixed saved indices before the physical field
decays. TangentPDE-Core better preserves the localized structures and reduces
the dominant error on this median-improvement trajectory.}
\label{fig:app_diffreact_temporal}
\end{figure*}

\clearpage
\begin{figure*}[p]
\centering
\includegraphics[width=0.94\textwidth]{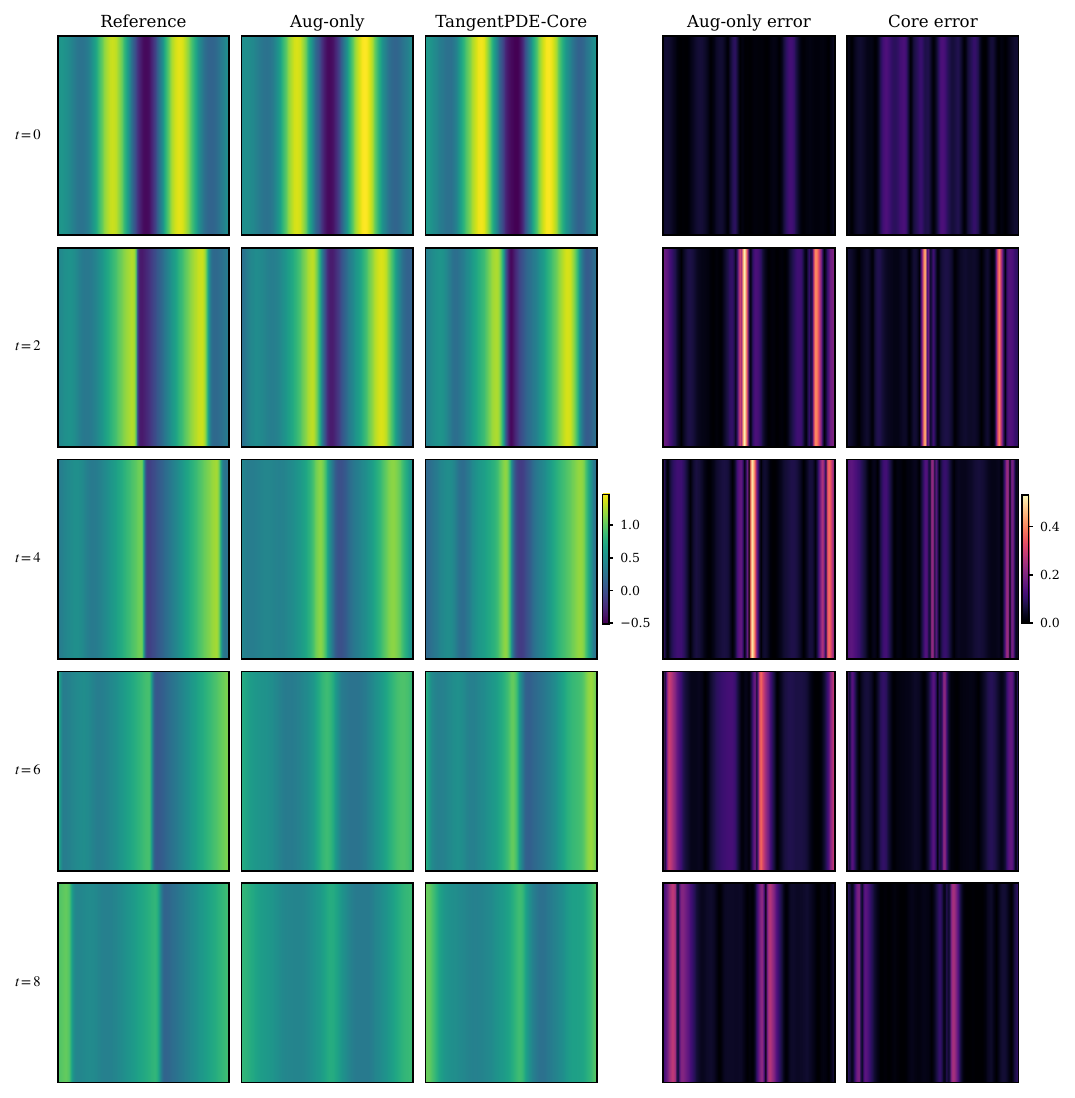}
\caption{Early-time Burgers one-dimensional state fields at saved rollout
indices $0,2,4,6,8$ from the independent uniform-protocol FNO rerun.
TangentPDE-Core more closely follows the transported structures and reduces
the dominant errors on this median-improvement trajectory. The color strips
are expanded vertically for readability.}
\label{fig:app_burgers_temporal}
\end{figure*}

\clearpage
\begin{figure*}[p]
\centering
\includegraphics[width=0.94\textwidth]{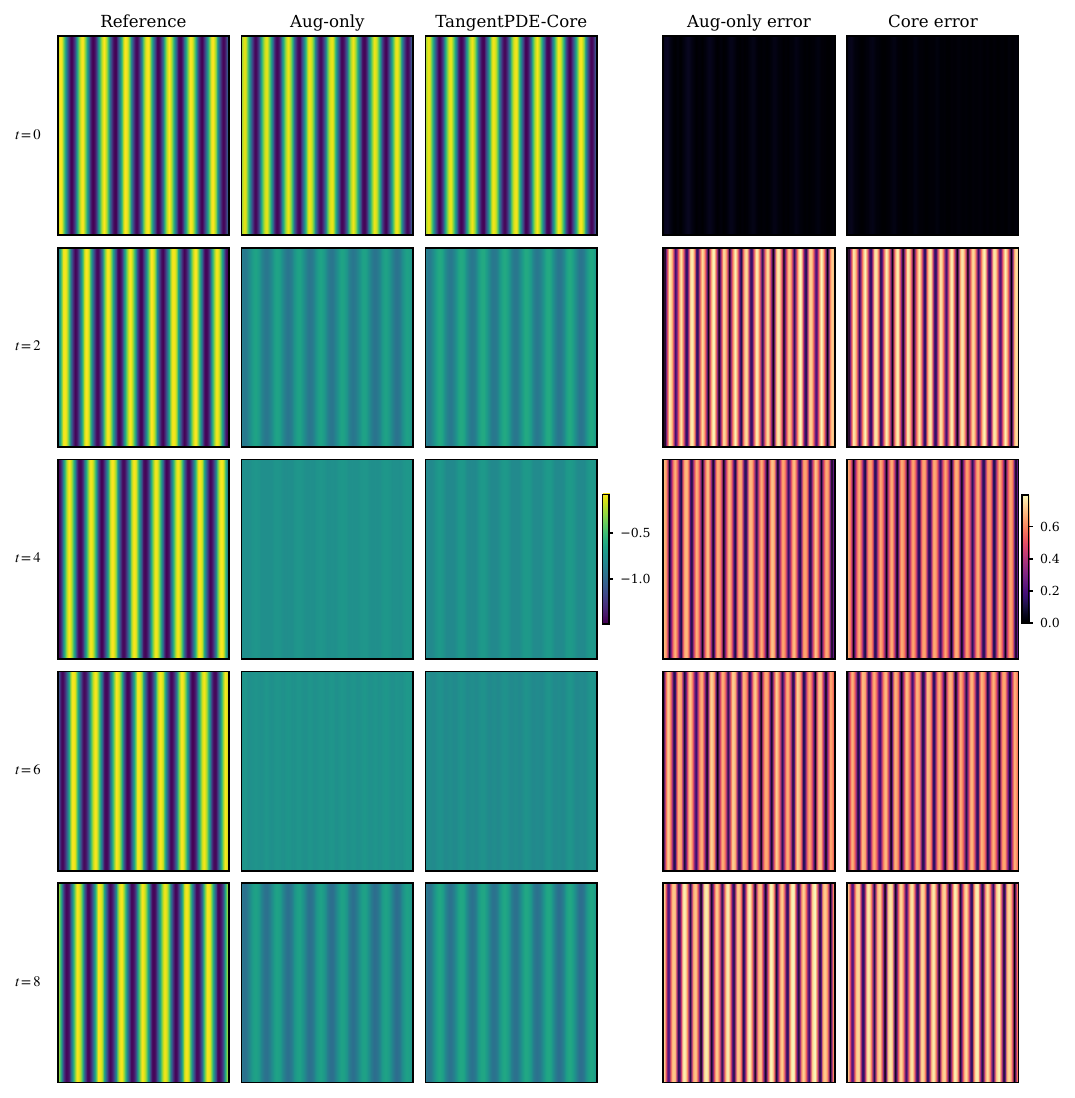}
\caption{Early-time Advection one-dimensional state fields at saved rollout
indices $0,2,4,6,8$ from the independent uniform-protocol FNO rerun. Aug-only
and TangentPDE-Core remain visually similar, consistent with the small and
mixed transport-task differences in the quantitative results. The color strips
are expanded vertically for readability.}
\label{fig:app_advection_temporal}
\end{figure*}

\FloatBarrier

\end{document}